\documentclass[11pt]{style/prism-princeton}

\makeatletter
\newcommand{\longdash}[1][2em]{%
  \makebox[#1]{$\m@th\smash-\mkern-7mu\cleaders\hbox{$\mkern-2mu\smash-\mkern-2mu$}\hfill\mkern-7mu\smash-$}}
\makeatother
\newcommand{\omitskip}{\kern-\arraycolsep}

\usepackage{amsmath,amssymb,stmaryrd,mathtools}
\usepackage{xcolor}
\usepackage{xspace}
\usepackage{bbm}
\usepackage{graphicx}
\usepackage{trimclip}

\newcommand{\iosemoji}[1]{%
  \raisebox{-0.12em}{\includegraphics[height=0.85em]{#1}}%
}

\usepackage{subfig}
\usepackage{tcolorbox}
\usepackage{colortbl}
\usepackage{multirow}
\usepackage{booktabs}
\usepackage{wrapfig}
\usepackage{lipsum}
\usepackage{cleveref}
\usepackage{twemojis}
\usepackage{caption}

\definecolor{stdgray}{gray}{0.45}

\usepackage{tabularx}
\usepackage{array}

\definecolor{dark_red}{RGB}{122, 0, 0}
\definecolor{coral}{RGB}{255, 119, 94}
\definecolor{pink_orange}{RGB}{255, 72, 126}
\definecolor{vibrant_pink}{RGB}{255, 0, 104}
\definecolor{pink_pink}{RGB}{255, 37, 153}
\definecolor{wine}{RGB}{204, 0, 102}
\definecolor{tango_red}{RGB}{172,12,45}

\definecolor{light_orange}{RGB}{255, 198, 107}
\definecolor{orange(sae/ece)}{rgb}{1.0, 0.49, 0.0}
\definecolor{dark_orange}{RGB}{216,92,0}

\definecolor{org-purp-0}{RGB}{165, 76, 0}
\definecolor{org-purp-1}{RGB}{250, 130, 28}
\definecolor{org-purp-2}{RGB}{226, 89, 68}
\definecolor{org-purp-3}{RGB}{206, 92, 124}
\definecolor{org-purp-4}{RGB}{116, 80, 146}
\definecolor{org-purp-5}{RGB}{110, 78, 157}

\definecolor{teal(sae/ece)}{rgb}{0, 0.47, 0.52}
\definecolor{aqua}{RGB}{52,172,139}
\definecolor{dark_aqua}{RGB}{35,115,93}
\definecolor{dark_green}{RGB}{0, 92, 34}

\definecolor{grape}{RGB}{112,48,160}
\definecolor{purple}{rgb}{0.74, 0.65, 1.0}
\definecolor{dark_purple}{rgb}{0.58, 0.0, 0.82}
\definecolor{periwinkle}{RGB}{191, 140, 230}

\definecolor{light_gray}{rgb}{0.9, 0.9, 0.9}
\definecolor{medium_gray}{rgb}{0.6, 0.6, 0.6} 
\definecolor{dark_gray}{rgb}{0.2, 0.2, 0.2} 

\definecolor{sky_blue}{RGB}{37, 166, 213}
\definecolor{light_blue}{rgb}{0.33, 0.80, 1}
\definecolor{dark_blue}{rgb}{0.098, 0.239, 0.52}
\definecolor{ocean}{RGB}{13, 121, 202}
\definecolor{light_ocean}{RGB}{18, 178, 235}
\definecolor{dark_ocean}{RGB}{10, 89, 148}
\definecolor{vibrant_blue}{RGB}{14, 120, 255}

\definecolor{dark_brown}{rgb}{0.3255, 0.004, 0.001}

\newcounter{qnum}
\newcounter{tnum}
\makeatletter
\@ifundefined{definition}{}{}
\makeatother

\newcommand{\ours}{\textcolor{tango_red}{\fontfamily{lmtt}\selectfont\textbf{TANGO}}\xspace}
\newcommand{\oursb}{\textcolor{black}{\fontfamily{lmtt}\selectfont\textbf{TANGO}}\xspace}

\usepackage{amsmath}

\usepackage{pifont}

\providecommand{\cmark}{\ding{51}}
\providecommand{\xmark}{\ding{55}}

\usepackage{wrapfig}

\newif\ifreviewchanges
\reviewchangesfalse
\newcommand{\reviewstyle}{%
  \color{dark_green}%
  \hypersetup{linkcolor=dark_green,citecolor=dark_green,urlcolor=dark_green}%
  \renewcommand{\ours}{{\fontfamily{lmtt}\selectfont\textbf{TANGO}}\xspace}%
  \let\oursb\ours
}
\DeclareRobustCommand{\reviewtext}[1]{%
  \ifreviewchanges{\reviewstyle #1}\else#1\fi}
\newenvironment{reviewblock}{\begingroup\ifreviewchanges\reviewstyle\fi}{\endgroup}

\makeatletter
\def\fps@table{t}
\makeatother

\author[1,2,*]{Anqi Li}
\author[1,3,*,\ddagger]{Yuxin Chen}
\author[1,*]{Zhaobo Li}
\author[1,4,*]{Zhuo Cao}
\author[1,5]{Junli Ren}
\author[1]{Masayoshi Tomizuka}
\author[6,\dagger]{Dhruv Shah}

\affiliation[1]{University of California, Berkeley}
\affiliation[2]{Peking University}
\affiliation[3]{NVIDIA}
\affiliation[4]{Tsinghua University}
\affiliation[5]{The University of Hong Kong}
\affiliation[6]{Princeton University}
\contribution[*]{Equal contribution.}
\contribution[\ddagger]{Project lead.}
\contribution[\dagger]{Corresponding author.}

\begin{document}

\title{\textcolor{tango_red}{\fontfamily{lmtt}\selectfont\textbf{TANGO}}\iosemoji{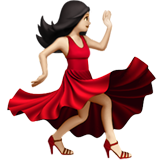}: Humanoid Navigation in Cluttered Environments with a Whole-Body Vision-Language-Action Model}

\abstract{
    We study the problem of navigating cluttered indoor environments with a humanoid robot.
Unlike conventional methods that \reviewtext{model} navigation as a 2D path planning problem, humanoid traversal in cluttered \reviewtext{environments} requires continuous geometry-aware whole-body adaptation, including coordinated arm placement, torso adjustment, and gait modulation for collision-free movement through complex 3D spaces.
We introduce \oursb, the first whole-body vision-language navigation framework for language-conditioned humanoid traversal in cluttered environments. 
Given a natural-language instruction and egocentric RGB observations, \oursb directly predicts 29-DoF joint-space actions for downstream whole-body control. 
We train \oursb entirely in simulation by synthesizing diverse collision-free traversal behaviors via global path planning, kinematic whole-body motion generation, obstacle-aware motion editing, and RL-based tracking. 
This pipeline provides dynamically feasible action supervision for learning language-conditioned whole-body policies. 
In extensive simulation experiments, \oursb demonstrates state-of-the-art performance in vision-language navigation, while outperforming strong modular baselines in navigating challenging scenes requiring obstacle negotiation. 
Lastly, we deploy \oursb zero-shot on a Unitree G1 humanoid robot, and observe robust language-guided traversal in cluttered real-world scenes without training on any real-world navigation data. 

}

\keywords{Vision-Language Navigation, Vision-Language-Action Model, Whole-Body Control}

\website{https://tango-vla.github.io/}{tango-vla.github.io}

\providecommand{\codelist}{}

\patchcmd{\mymaketitle}
  {\websitelist\par}
  {\websitelist\par\vspace{0.1cm}\newcommand{\logogap}[1]{\hfill\kern#1\linewidth\relax}
\noindent\makebox[\linewidth]{%
  \raisebox{-0.5\height}{\includegraphics[trim=21bp 596bp 21bp 596bp,clip,width=0.15\linewidth]{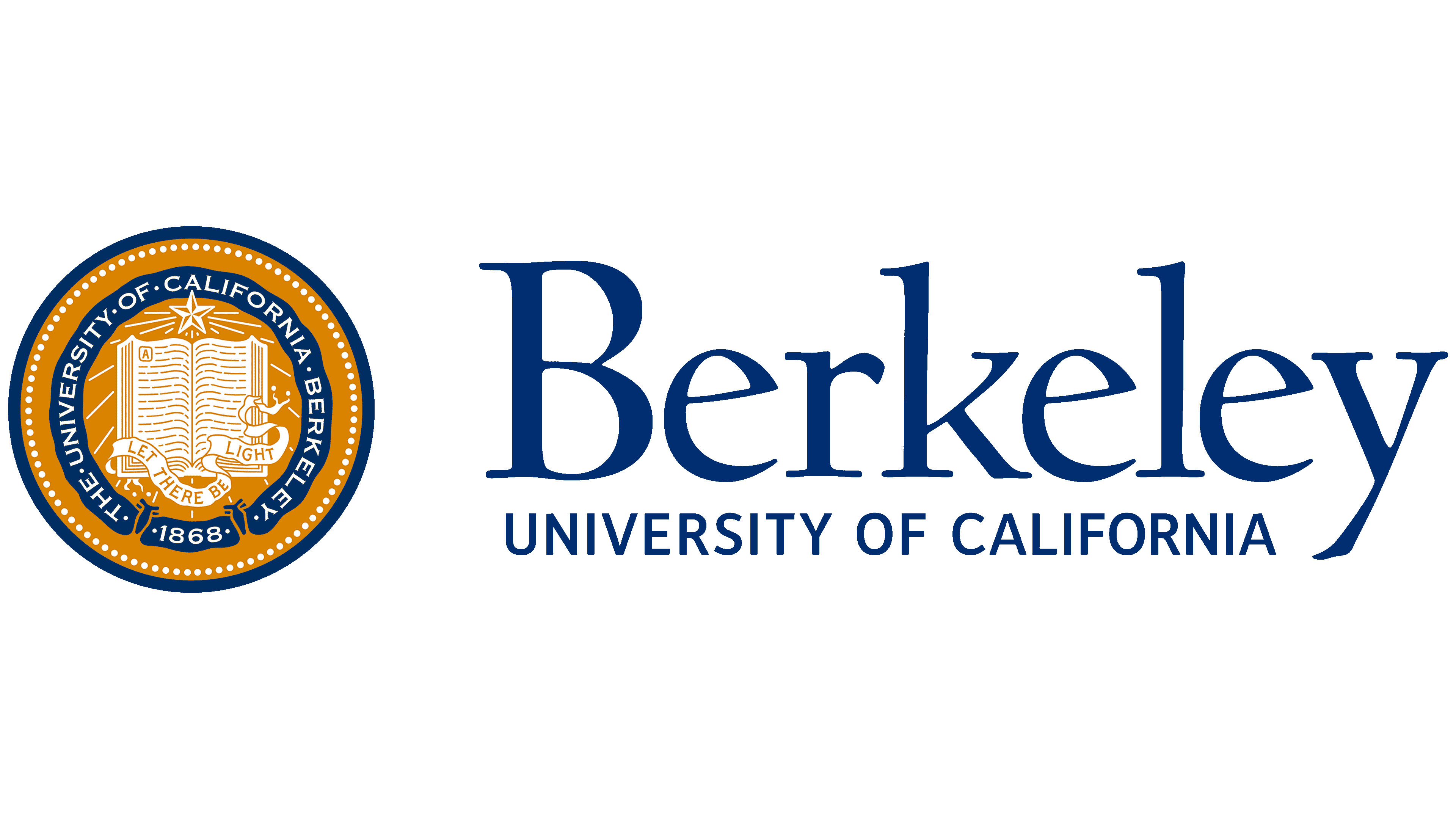}}%
  \logogap{-0.0002}%
  \raisebox{-0.5\height}{\includegraphics[width=0.14\linewidth]{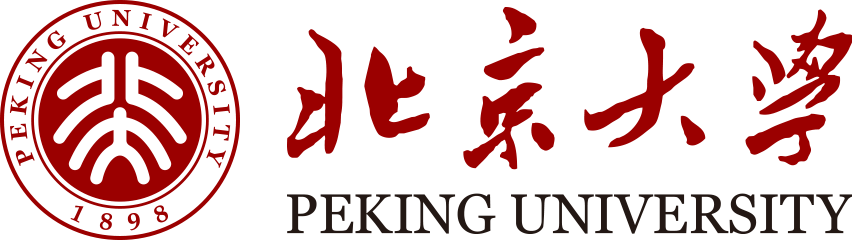}}%
  \logogap{-0.0007}%
  \raisebox{-0.5\height}{\includegraphics[trim=0bp 0bp 18.8bp 0bp,width=0.14\linewidth]{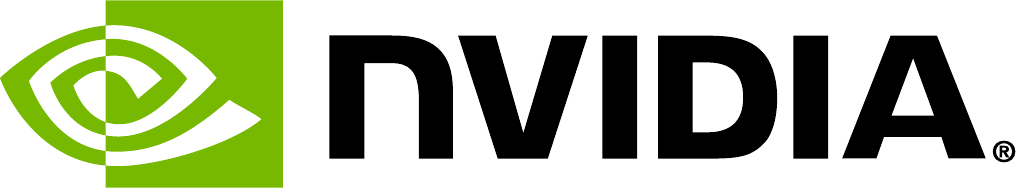}}%
  \logogap{0.0002}%
  \raisebox{-0.5\height}{\includegraphics[width=0.13\linewidth]{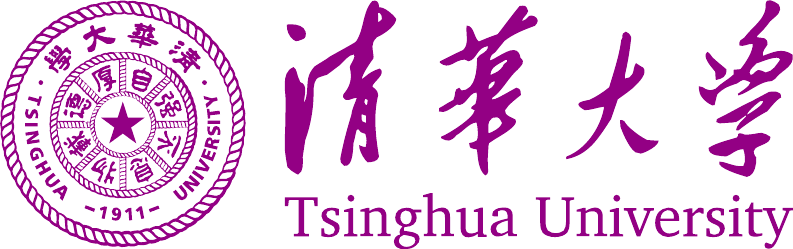}}%
  \logogap{0.0014}%
  \raisebox{-0.5\height}{\includegraphics[width=0.175\linewidth]{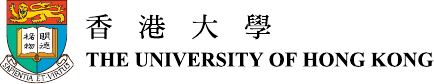}}%
  \logogap{-0.0007}%
  \raisebox{-0.5\height}{\includegraphics[trim=16bp 10bp 3bp 19bp,width=0.16\linewidth]{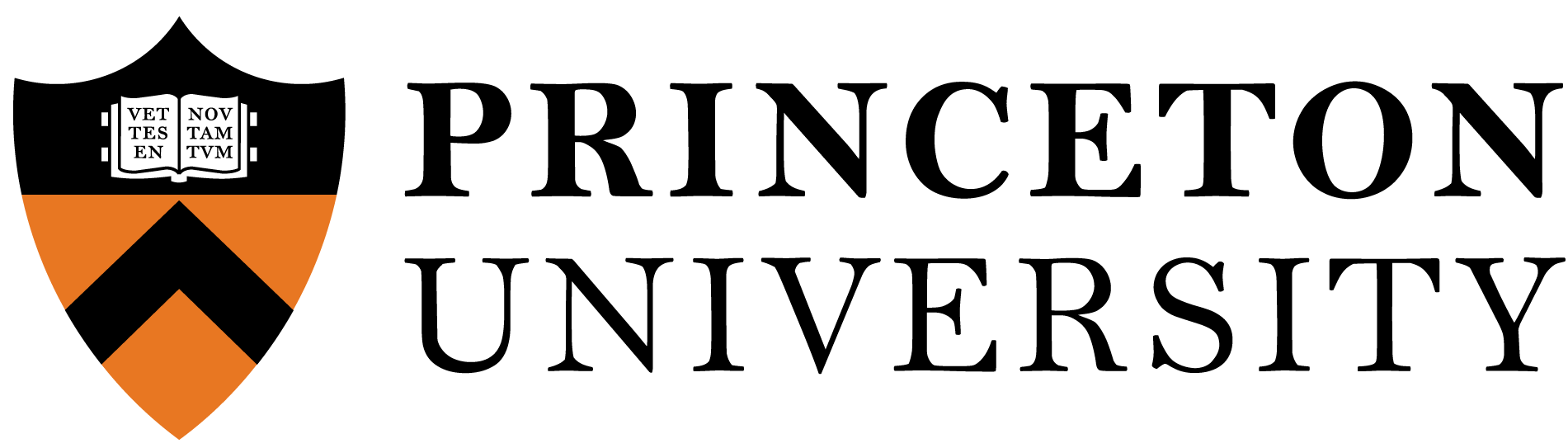}}%
}\par
}
  {}{\PackageError{tango}{Could not insert affiliation logos}{Check the title template.}}

\makeatletter
\let\tango@maketitle\maketitle
\renewcommand{\maketitle}{%
    \begingroup
    \let\tango@includegraphics\includegraphics
    \renewcommand{\includegraphics}[2][]{%
        \ifnum\pdfstrcmp{##2}{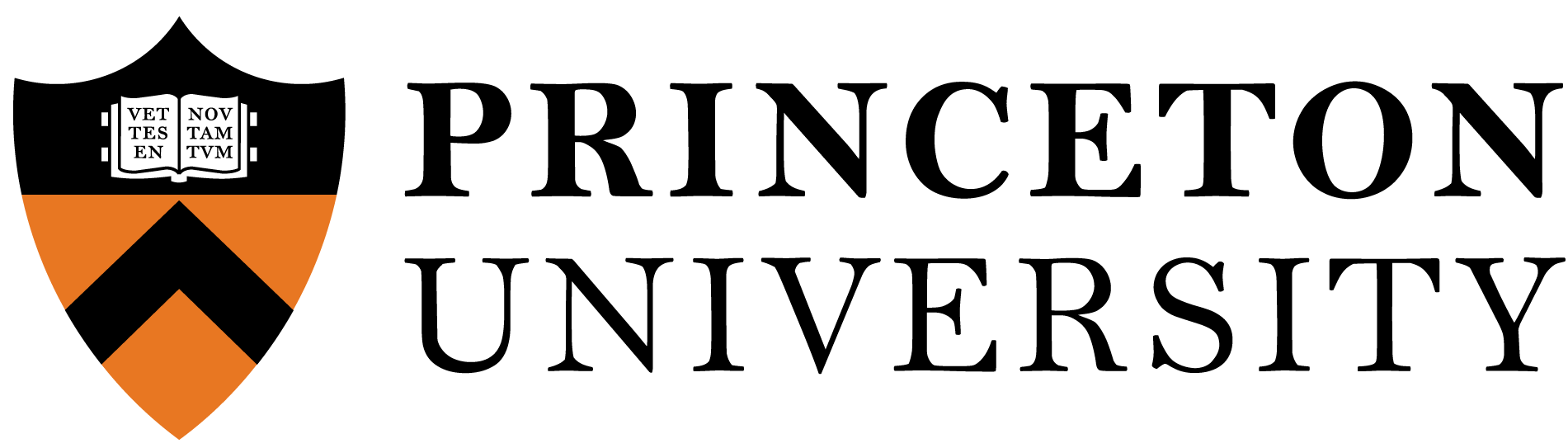}=0
        \else
            \tango@includegraphics[##1]{##2}%
        \fi
    }%
    \tango@maketitle
    \endgroup
}
\makeatother

\maketitle

\begin{figure}[t]
    \centering
    \includegraphics[width=\textwidth]{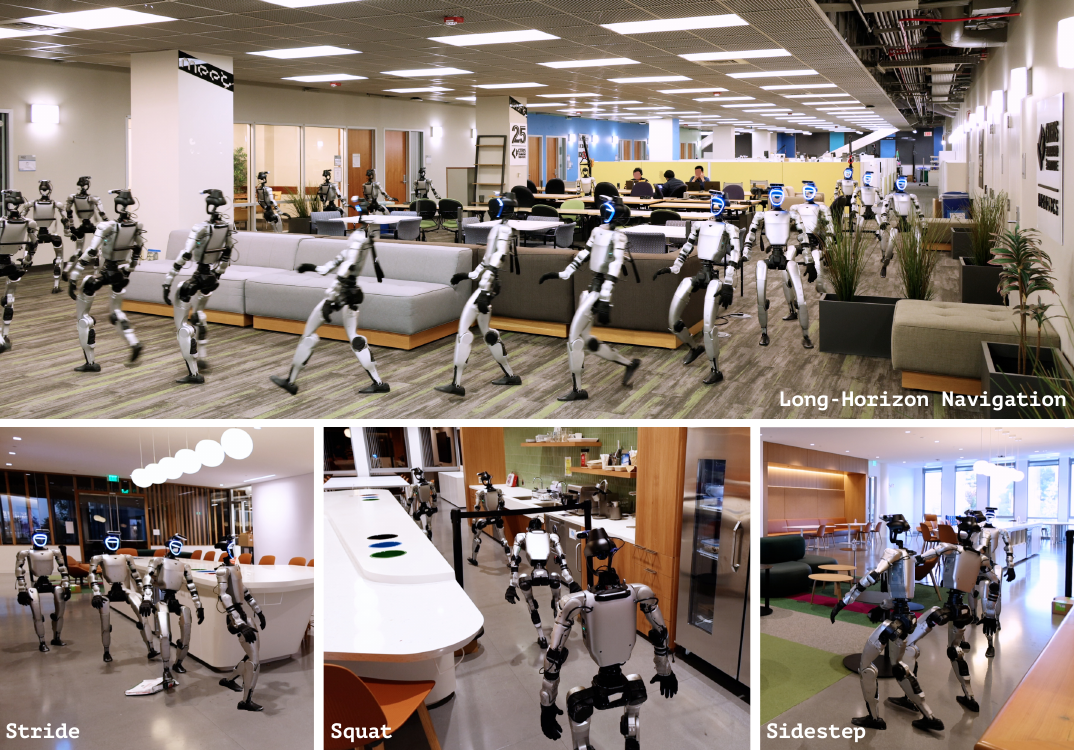}
    \caption{\textbf{Humanoid whole-body navigation in cluttered environments.} We propose \oursb, a whole-body foundation model for cluttered indoor scene navigation. \oursb demonstrates strong scene understanding and robust traversal capability, navigating a 30-meter route zero-shot in a real-world cluttered office scene, and traversing through cluttered scenes with overhead and floor obstacles, as well as narrow passages. }
    \label{fig:teaser}
\end{figure}

\section{Introduction}
Language-guided traversal in complex 3D environments is a fundamental capability for domestic humanoid robots expected to assist with everyday tasks.~\cite{gu2026humanoid}
Unlike wheeled or mobile-base robots, humanoids navigate with high-dimensional articulated bodies whose geometry changes continuously during motion, making the robot’s body configuration an inherent part of the navigation problem. 
As illustrated in~\Cref{fig:teaser}, traversal feasibility in cluttered indoor spaces depends not only on the intended route but also on whether the robot can physically move through the surrounding scene geometry without collisions involving the arms, torso, or legs. 
This creates a tight coupling between navigation decisions and whole-body feasibility: \textit{An action that appears valid at the planning level may still be infeasible for the embodied humanoid to execute.} 
Despite rapid progress in humanoid control~\cite{luo2025sonic, ze2025twist2, liao2025beyondmimic, videomimic}, vision-language navigation (VLN)~\cite{anderson2018vln, zhang2025embodied, zhang2024navid, hirose24lelan, wei2025internvln, cheng2025navila}, and collision-aware traversal~\cite{wei2025internvln,xue2026cat,xu2025mmnav}, effectively integrating these capabilities remains largely unexplored.

Existing VLN methods typically formulate navigation as high-level decision making, where an agent predicts 2D waypoints or discrete actions from visual observations and language instructions~\cite{shah2023vint,shah2023lmnav,hirose2025omnivla}. While effective for mobile platforms and simplified embodied agents, their low-dimensional action spaces cannot explicitly represent the relationship between navigation decisions and whole-body feasibility, limiting their ability to handle spatially constrained traversal scenarios.
Recent humanoid foundation models and whole-body VLA systems~\cite{wei2026psi0,gear2025grootn16,jiang2025wholebodyvla, ding2025humanoidvla} have demonstrated impressive whole-body control capabilities. Nevertheless, navigation in these systems is typically represented through high-level locomotion commands and delegated to downstream controllers, preventing explicit reasoning about whole-body traversability during navigation.
A complementary line of work explores collision-aware humanoid traversal through reinforcement learning~\cite{xue2026cat,wu2026perceptivehumanoidparkourchaining}. Although effective in specific traversal scenarios, these approaches often rely on task-specific priors or training distributions, limiting their scalability to long-horizon language-guided navigation in diverse cluttered environments.
Consequently, existing approaches remain unable to jointly reason about navigation intent and whole-body traversability, motivating the need for a unified framework for language-guided whole-body navigation.

To address these challenges, we present
\textbf{\color{tango_red}T}raversability-\textbf{\color{tango_red}A}ware Vision-Language \textbf{\color{tango_red}N}avi\textbf{\color{tango_red}g}ati\textbf{\color{tango_red}o}n
(\ours), a unified VLA framework for humanoid navigation in cluttered environments. Given a language instruction and egocentric RGB observations, \oursb directly predicts 29-DoF joint-space actions for whole-body humanoid control, avoiding the need for separate navigation and control modules.
A key challenge is obtaining large-scale training data that captures both semantic task diversity and physically plausible whole-body traversal behaviors. To this end, we develop a scalable simulation pipeline that automatically synthesizes collision-free humanoid traversal trajectories and provides dynamically feasible supervision for training the VLA model.
At deployment, the learned policy is executed through a robust motion tracker with real-time action chunking~\cite{luo2025sonic}, enabling reliable execution and zero-shot transfer to real humanoid hardware. We evaluate \oursb against state-of-the-art VLN and humanoid spatial traversal baselines in both simulation and real-world environments. Across long-horizon navigation tasks requiring obstacle negotiation and geometry-aware whole-body adaptation, \oursb consistently achieves stronger collision-free traversal performance than existing methods. We will open-source the data pipeline, generated dataset, VLA framework, model checkpoint, and deployment system to facilitate reproducibility.

\section{Related Works}
\textbf{Large Models for Vision-Language Navigation. }
Recent large multi-modality models (LMMs) emerge with strong scene understanding and physical awareness, leading to extensive zero-shot navigation works that leverage off-the-shelf large models~\cite{zhou2023navgpt, zhou2024navgpt2, chandaka2025reasonnav, rajabi2025travel}.
Moreover, recent efforts have explored fine-tuning such models on simulated and real-world navigation samples, resulting in strong VLA models for navigating in diversified environments~\cite{zhang2024navid,cheng2025navila,li2025urbanvla,wei2025internvln, zhang2025embodied, hirose2025omnivla}. Nevertheless, these methods take visual navigation as a pure planar trajectory planning task, omitting the physical gap~\cite{wang2025vlnpe} when deployed in real physical environments. In contrast, \oursb is trained with inherent physical awareness, mitigating the embodied gap while enabling explicit reasoning about whole-body traversability in cluttered environments. 
Recent efforts have also explored dual-system design for VLA models to achieve continuous, real-time navigation~\cite{wei2025internvln, hirose2026asyncvla}. In this work, we equip \oursb with a flow-matching-based action expert as \textit{system-1}, trained with real-time chunking~\cite{black2025trainingtimertc}, to achieve continuous and responsive humanoid control.

\textbf{Cluttered Environment Traversal. }
Traversal in cluttered scenes is critical for deploying embodied agents in complex real-world scenarios. Recent humanoid parkour works have demonstrated impressive traversal capabilities over challenging terrains and obstacles~\cite{zhuang2024humanoidparkourlearning, wu2026perceptivehumanoidparkourchaining, ASTraversal}. However, these methods mainly focus on short-horizon interactions with scene objects. In contrast, \oursb performs long-horizon navigation with collision avoidance, which requires excellence in physical and semantic understanding capabilities. 
HumanoidPF~\cite{xue2026cat} introduces RL-based collision-free indoor traversal for humanoids and achieves high success rates in most cases, but remains difficult to scale, especially to long-horizon navigation and complex obstacle compositions. In this work, we synthesize collision-free and dynamically feasible humanoid motions through a scalable pipeline, collecting low-cost, high-quality datasets for 3D traversal policy training. Some VLN works~\cite{xu2025mmnav,liu2026spannav} also study traversal in cluttered scenes, but are fundamentally limited by their 2D problem formulation and primarily consider bypassing behaviors. In contrast, \oursb learns humanoid whole-body motions, enabling richer capabilities when facing complex obstacles.

\textbf{Humanoid Whole-Body Control through Large-Scale Learning. }
Recent advances in humanoid motion tracking~\cite{li2025amo, ze2025twist2, luo2025sonic} have enabled large-scale learning of humanoid control policies. Representative works such as GR00T-N1.6~\cite{gear2025grootn16}, $\Psi_0$~\cite{wei2026psi0}, and WholeBodyVLA~\cite{jiang2025wholebodyvla} adopt a decoupled design, predicting upper-body motions while issuing high-level commands to a lower-body tracker. While this significantly simplifies loco-manipulation learning, it limits whole-body coordination required for tasks such as cluttered-scene traversal. In contrast, \oursb directly learns end-to-end whole-body motions and uses them as reference trajectories for a low-level tracker. Non-decoupled approaches, including LeVERB~\cite{xue2025leverb}, HumanoidVLA~\cite{ding2025humanoidvla}, and PhysiFlow~\cite{qin2026physiflow}, learn latent motion representations decoded by specialized controllers. Instead, \oursb directly predicts executable whole-body actions and relies on a pre-trained general-purpose tracker for execution, avoiding controller co-training and task-specific motion decoders while enabling scalable deployment across diverse humanoid platforms and traversal tasks.

\section{Method}
We present \oursb, a whole-body VLA system for cluttered indoor vision-language navigation (\Cref{fig:pipeline}).
In this section, we first provide a formal definition of whole-body vision-language navigation (\Cref{sec:WBVLN}).
Next, we introduce an automated data-generation pipeline for building a large-scale whole-body navigation dataset through scalable scene augmentation and motion editing in simulation environments (\Cref{sec:data_augmentation}).
We then describe how \oursb generates whole-body motions from language instructions and RGB observations (\Cref{sec:main_method}).
Lastly, we describe how to deploy \oursb both in simulated environments and on a real humanoid robot (\Cref{sec:deployment}).

\subsection{Problem Formulation}
\label{sec:WBVLN}
Existing VLN works are fundamentally limited by their planar action spaces, failing to represent complex traversing motions in real-world environments.
In this work, we study the problem of \textbf{\textit{whole-body vision-language navigation}}.
Given a natural language instruction $\ell$, current observation $\mathbf{o}_t$ containing a temporal sequence of RGB images from front and downward cameras $\mathbf{I}_{1:t}^{\text{fr}, \text{dn}}$ and whole-body joint-angle proprioceptive state $\mathbf{q}_t$, 
our model learns to predict a \emph{whole-body action chunk} $\mathbf{A}_t = \{\mathbf{a}_1, \cdots, \mathbf{a}_{H}\}$ over an action horizon $H$, 
where $\mathbf{a}_i = \{\mathbf{q}_{\text{d},i}, \mathbf{r}_{\text{b},i}\},~i\in\{1, \cdots, H\}$,
with $\mathbf{q}_{\text{d},i} \in \mathbb{R}^{29}$ and $\mathbf{r}_{\text{b},i} \in \mathbb{R}^{6}$ denoting the desired whole-body joint angles and the base 6D rotation representation at the $i$-th step, respectively.
The predicted action chunk is then streamed to a low-level motion tracker for physically grounded humanoid navigation.

\subsection{Simulation Data Generation}
\label{sec:data_augmentation}
\begin{figure*}[t]
  \centering
  \includegraphics[width=0.9\textwidth]{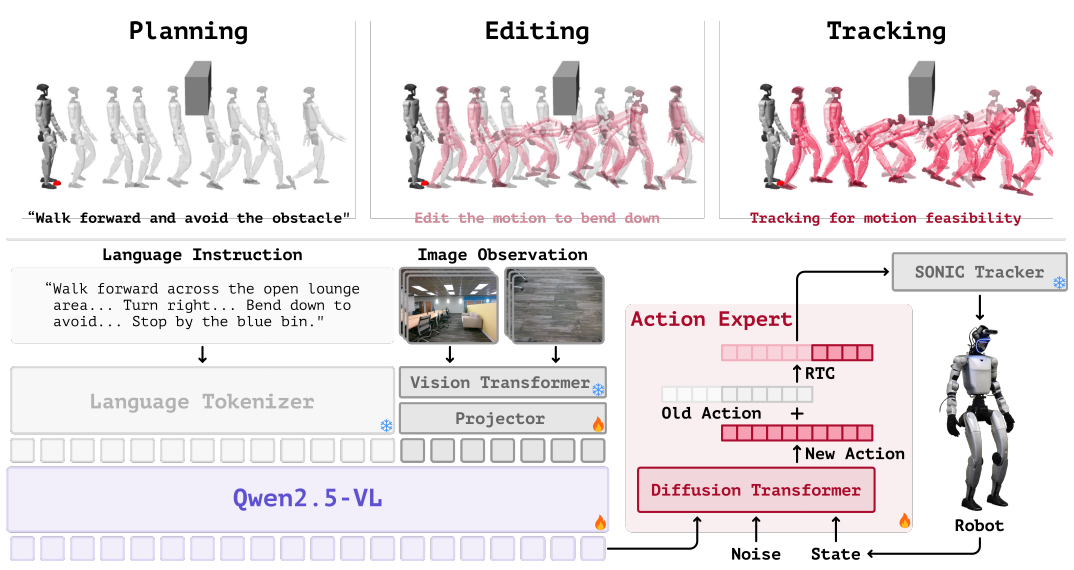}
  \caption{
    \textbf{\oursb Architecture and Data Pipeline. }
    Top: the proposed Plan--Edit--Track (PET) pipeline automatically synthesizes collision-free whole-body traversal data. Bottom: \oursb combines a vision-language backbone, a diffusion-based action expert, and a low-level tracker for language-guided whole-body humanoid control.
  }
  \label{fig:pipeline}
  \vspace{-10pt}
\end{figure*}

\textbf{Environment Augmentation. }
Existing navigation datasets~\cite{anderson2018vln, lin2025vlnverse, miao2025sage3d} primarily capture standard room layouts
, while real-world environments are characterized by randomly placed objects that create complex spatial constraints for robot traversal.
To create more challenging scenarios, we augment indoor scenes from \textit{VLNVerse}~\cite{lin2025vlnverse} and \textit{SAGE-3D}~\cite{miao2025sage3d}, which originally contain 263 and 1,000 scenes, respectively. After filtering low-quality scenes using Gemini 2.5 Flash, we retain 205 \textit{VLNVerse} scenes and 373 \textit{SAGE-3D} scenes, totaling 578 source scenes for augmentation.
Following HumanoidPF~\cite{xue2026cat}, we introduce three obstacle categories: lateral obstacles to construct narrow passages, ground-level obstacles to necessitate stepping, and overhead obstacles to enforce upper-body clearance. To ensure visual and semantic consistency, we curate category-specific templates from existing assets within the source datasets and automatically instantiate these semantically matched objects via $\mathrm{SO}(3)$ transformations. Please refer to~\Cref{app:env_aug_details} for further details.

\textbf{Collision-Free Motion Generation. }Humanoid motion collection pipelines often rely on labor-intensive human motion capture followed by retargeting~\cite{he2024learning, fu2025humanplus, ze2025twist}, which introduces motion degradation and embodiment mismatch. To generate scalable, human-like traversal data, we propose \textit{Plan, Edit, Track} (PET), an automatic pipeline that synthesizes collision-free, whole-body motions for humanoid cluttered indoor navigation, in an offline manner. Given limited space, full details are discussed in~\Cref{app:edit}.

Given a start and goal location in an augmented scene, PET first \textit{plans} a collision-aware planar reference path using A* with an obstacle-aware \reviewtext{path cost} that biases the search toward safer regions. A heading adjustment module detects narrow passages and inserts 90-degree heading changes, inducing sideways walking when frontal traversal is spatially constrained. The resulting trajectory is converted into velocity commands for the SONIC motion planner~\cite{luo2025sonic} to produce natural humanoid walking motions along the planned 2D trajectories. We then replay these trajectories in IsaacSim~\cite{isaacsim} to render egocentric RGB observations at 2 Hz, which is then fed into Gemini 2.5 Flash~\cite{comanici2025gemini25} to generate formatted VLN instructions, following previous work~\cite{lin2025vlnverse}.

PET subsequently \textit{edits} the synthesized motions to incorporate whole-body obstacle interactions. We apply sampled guidance forces from humanoid potential field~\cite{xue2026cat} conducted with obstacle and ground truth trajectory priors to key body links through SoftMimic-style pseudo-forces~\cite{margolis2025softmimic}. To negotiate ground-level obstacles, our gait-adaptation module retargets foot landing positions beyond each obstacle and adjusts swing-foot clearance while preserving the reference gait phase and timing.
This stage yields reference motions with explicit whole-body avoidance behaviors, such as arm clearance, stepping over, and crouching.

Finally, PET employs a SONIC \textit{tracker} as an executability filter. The edited motions are tracked in simulation to verify physical feasibility and collision freedom. Failing or colliding trajectories are discarded. Crucially, instead of using the tracked trajectories as training supervision, which might degrade the human-like quality, we utilize the collision-free reference motions from the planning and editing stages as the action supervision signal. This preserves human-like structures while ensuring physical executability. Based on the verified trajectories, we regenerate the RGB observations and update the language instructions accordingly. The resulting dataset contains 64,633 trajectories; PET and rendering require 86 and 125 RTX PRO 6000 GPU-hours, respectively, for a total of 211 GPU-hours (\Cref{tab:data_generation_cost}).

\subsection[TANGO Architecture and Training]{\oursb Architecture and Training}
\label{sec:main_method}
Whole-body VLA navigation in complex environments demands physical world understanding, continuous action prediction, and robust action execution.
To address these requirements, \oursb adopts a \textit{triple-system} architecture~\cite{intelligence2025pi05visionlanguageactionmodelopenworld, nvidia2025gr00tn1openfoundation, wei2026psi0} integrating a Vision-Language (VL) backbone (\textit{system-2}), a multi-modal diffusion transformer (MM-DiT) action expert with real-time chunking~\cite{black2025trainingtimertc} (\textit{system-1}), and an off-the-shelf motion tracker (\textit{system-0}), as shown in~\Cref{fig:pipeline}. 
We jointly train the VL backbone and the action expert. During deployment, the predicted action chunks are streamed to the low-level tracker to generate continuous, high-frequency humanoid control signals.

\textbf{System-2: Vision-Language Perception.} 
We instantiate \textit{system-2} using Qwen2.5VL-7B~\cite{bai2025qwen25vl}, warm-started with InternVLA-N1~\cite{wei2025internvln} weights to inherit strong navigation priors. We choose InternVLA-N1 for its open-source availability and navigation pretraining, while \reviewtext{other} VLA-based VLN backbones could also be adapted to this framework.
At timestep $t$, front and downward camera views ($\mathbf{I}_{t}^{\text{fr}}, \mathbf{I}_{t}^{\text{dn}}$) are vertically stacked into a single frame $\mathbf{I}_{t}$. 
To manage long-horizon video history $\mathbf{I}_{1:T}$ within a given memory capacity, we apply Budget-Aware Token Sampling (BATS)~\cite{zhang2025embodied}. 
In the time step $T$, all history frames are sampled independently into the navigation context according to the probability function
$
P(t) = (1-\epsilon)e^{k(t-T)/T} + \epsilon,\ t \in [1,T],
$
where $\epsilon$ and $k$ regulate temporal intensity. The visual features $v_i = \text{VisionEncoder}(\mathbf{I}_i)\in \mathbb{R}^{n\times p\times c}$ are further spatial-grid pooled via $\tilde v_i = \mathrm{GridPool}(v_i, g_i)\in\mathbb{R}^{g_i\times c}$~\cite{zhang2024uninavid}, allocating finer grids to recent observations and coarser grids to history. Finally, the sampled visual tokens $\tilde v_{\text{sampled}}$ and language instruction $\ell$ are fed into the VLM to produce a latent context token $z = \text{VLM}(\tilde v_{\text{sampled}}, \ell)$.

\textbf{System-1 \& System-0: Action Prediction and Execution.}
Conditioned on the latent $z$ and current proprioception $\mathbf{q}_{d,t}$, the \textit{system-1} action expert predicts a future whole-body reference chunk.
While standard actions are defined as joint angles and base poses $\mathbf{a}_i=\{\mathbf{q}_{d,i},\mathbf{r}_{b,i}\}$, we formulate a stabilized training target to facilitate regression
$
\tilde{\mathbf{a}}_i=\{\mathbf{q}_{d,i},\tilde{\mathbf{r}}_{b,i},\Delta x_i,\Delta y_i,\Delta\psi_i\},
$
where the base yaw in $\tilde{\mathbf{r}}_{b,i}$ is parameterized relative to the first frame of the chunk, and the auxiliary deltas ($\Delta x_i,\Delta y_i,\Delta\psi_i$) explicitly encode chunk-level planar displacement and heading changes.
We implement \textit{system-1} using a flow-based MM-DiT~\cite{esser2024mmdit} trained via flow-matching to generate the horizon $\tilde{\mathbf{A}}_{t:t+H}$. To align offline training with online streaming execution, we apply training-time RTC~\cite{black2025trainingtimertc}, conditioning the model on a randomized committed prefix of $d$ actions to inpaint the remaining horizon. The generated chunk is subsequently recovered to $\mathbf{A}_{t:t+H}$ and streamed to the SONIC tracker (\textit{system-0}), which tracks the reference against robot proprioception to provide high-frequency joint commands.

\textbf{Joint Training Objectives.} 
Following dual-branch designs for VL decoding~\cite{wang2025trackvla, zhang2025embodied}, we append a text-decoding branch to System-2 and co-tune navigation tasks alongside VideoQA samples~\cite{shen2024longvuspatiotemporaladaptivecompression} to preserve generalized world knowledge. 
The joint optimization objective is defined as $\mathcal{L} = \mathcal{L}_{\text{CE}} + w_{\text{FM}} \cdot \mathcal{L}_{\text{FM}}$, where $\mathcal{L}_{\text{CE}}$ is the cross-entropy loss for VideoQA, $\mathcal{L}_{\text{FM}}$ denotes the flow-matching loss, and $w_{\text{FM}} = 20$. \oursb is trained end-to-end for a single epoch with a learning rate of $1\times10^{-5}$.

\subsection{Deployment}
\label{sec:deployment}
We design the deployment system of \oursb on a Unitree G1 in both simulation and real world for evaluating the effectiveness and robustness of our system in both scenarios. 

\newpage
\begin{wrapfigure}{r}{0.48\textwidth}
\vspace{-3pt}
\centering

\includegraphics[width=\linewidth]{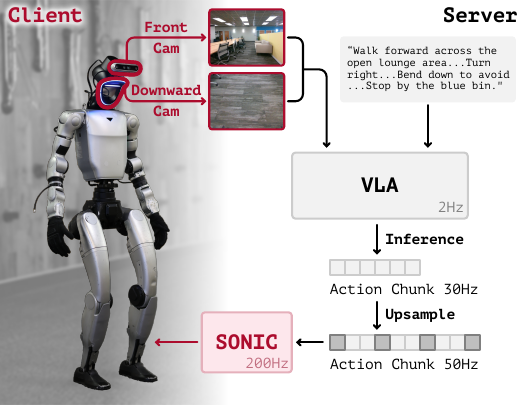}

\caption{
\textbf{\oursb real-world deployment system.} We adopt a server-client design to separately run low-frequency VLA model and high-frequency whole-body tracker.
}
\label{fig:deployment}

\end{wrapfigure}

\textbf{Simulation Deployment. }The task of whole-body navigation requires both high rendering quality and physical authenticity. To achieve both, we employ a digital twin teleportation system in simulation, where we leverage MuJoCo~\cite{mujoco} for low-level tracker deployment and physical simulation, and teleport a humanoid digital twin in IsaacSim~\cite{isaacsim} to acquire photo-realistic visual observation from designated camera pose solved from a given humanoid robot pose using forward kinematics (FK), and use it as VLA input.

\textbf{Real-World Deployment. }
We design a robust, plug-and-play real-world deployment system for \oursb. As shown in~\Cref{fig:deployment}, \oursb adopts a cloud-edge deployment architecture that separates the compute-intensive VLA module from the high-frequency WBC module. The VLA (\textit{system-2} and \textit{system-1}) runs on a cluster server equipped with an RTX PRO 6000, while the SONIC tracker (\textit{system-0}) runs on an onboard Jetson Orin NX. The two systems are connected through standard IP networking. The humanoid captures front- and downward-facing RGB observations using RealSense D455 and D435i cameras and streams them, together with proprioceptive states, to the server with approximately 20ms latency. The server serves as a global clock and performs VLA inference every 0.5s, matching an execution horizon of $s=15$ actions at 30Hz. Predicted motion chunks are streamed back to the robot, resampled to 50Hz, and executed by the SONIC tracker, which closes the low-level control loop at approximately 200Hz. Additionally, we design a webpage-based control \reviewtext{panel} with a user-friendly interface for sending navigation instructions, monitoring VLA output and robot observation, and sending control signals to the VLA and WBC systems. Please refer to~\Cref{app:impl_details} for more details.

\par\WFclear

\newlength{\experimentfloatsep}
\setlength{\experimentfloatsep}{\textfloatsep}
\setlength{\textfloatsep}{12pt plus 2pt minus 2pt}

\section{Experiments}
To evaluate the effectiveness of our method, we conduct extensive experiments and ablation studies to answer three key questions: 1) Can \ours perform well on VLN tasks compared to state-of-the-art baselines? 2) Can \ours effectively learn to traverse through cluttered indoor environment without collision? and 3) Is the key design of our method effective?

\suppressfloats[t]
\begin{table*}[t]
\centering
{\small

\clipbox{0pt 0pt 0pt 0pt}{%
\begin{tabular}{lccccccccc}

\toprule

\multirow{2}{*}{Methods}
& \multirow{2}{*}{\shortstack{Low-level\\Control}}
& \multicolumn{4}{c}{Val Seen}
& \multicolumn{4}{c}{Val Unseen}
\\

\cmidrule(r){3-6}
\cmidrule(r){7-10}

&
& NE$\downarrow$ & OSR$\uparrow$ & SR$\uparrow$ & SPL$\uparrow$
& NE$\downarrow$ & OSR$\uparrow$ & SR$\uparrow$ & SPL$\uparrow$
\\

\midrule

{CMA~\cite{krantz2020beyond}}
& \xmark
& 5.36 & 59.81 & 37.35 & 33.36
& 5.16 & 62.79 & 31.15 & 27.92
\\

RDP~\cite{wang2025vlnpe}
& \xmark
& {4.02} & \underline{68.09} & {47.28} & \textbf{41.69}
& \underline{3.75} & \textbf{71.93} & \underline{48.60} & \textbf{42.72}
\\

{Seq2Seq~\cite{krantz2020beyond}}
& \xmark
& 4.78 & 44.68 & 32.62 & 30.39
& 4.36 & 49.58 & 35.03 & 33.37
\\

HNR~\cite{Wang_lookahead} & \xmark
& -- & -- & 36.34 & 32.10 & -- & -- & 32.95 & 29.56 \\

InternVLA-N1~\cite{wei2025internvln} & \xmark
& {\underline{3.91}} & 62.50 & 51.56 & 34.37 & 4.09 & 64.13 & 45.56 & 34.98 \\

Uni-NaVid~\cite{zhang2024uninavid} & \xmark
& 4.31 & 60.62 & {\underline{51.88}} & 39.72 & 3.97 & 62.50 & 45.00 & 39.42 \\

\rowcolor{tango_red!25}[\tabcolsep][\dimexpr\tabcolsep+0.2pt\relax]
Ours
& \cmark
& \textbf{3.90} & \textbf{70.31} & \textbf{54.69} & \underline{40.18}
& \textbf{3.72} & \underline{71.07} & \textbf{52.89} & \underline{40.18}
\\

\bottomrule
\end{tabular}%
}
}

\caption{\textbf{\textit{VLNVerse} benchmark result. }We evaluate \ours on \textit{VLNVerse} benchmark against strong baselines with different action representations. Among all methods, only \ours is enabled with low-level physical control, while others are evaluated in a teleportation setting. {A dash denotes an unreported metric. Bold and underlined values mark the best and second-best results, respectively.} }
\label{tab:baseline}
\end{table*}

\subsection{VLN Performance}

\looseness=-1
We evaluate \ours on \textit{VLNVerse}, a newly established VLN benchmark with photorealistic indoor scenes in IsaacSim. We compare against strong baselines in the benchmark, including discrete action models CMA and Seq2Seq~\cite{krantz2020beyond}, continuous action model RDP~\cite{wang2025vlnpe}, neural implicit representation method HNR~\cite{Wang_lookahead}, and state-of-the-art VLA models InternVLA-N1~\cite{wei2025internvln} and Uni-NaVid~\cite{zhang2024uninavid}.
We evaluate all methods on the fine-grained validation splits. 
We report Success Rate (SR), Success weighted by Path Length (SPL), Navigation Error (NE), and Oracle Success Rate (OSR). To ensure fair comparison, all methods are trained on 3963 trajectories from \textit{VLNVerse-train} and evaluated on 423 and 825 trajectories from \textit{VLNVerse-seen} and \textit{VLNVerse-unseen}, respectively. For metric calculation details, please refer to \Cref{app:metrics}. We fine-tune InternVLA-N1 and Uni-NaVid on \textit{VLNVerse-train} for five epochs following their original training settings. Results for CMA, Seq2Seq, RDP, and HNR are provided by the \textit{VLNVerse} team.

\looseness=-1
Since all baselines lack low-level control modules, we evaluate them in a teleportation setting following the original \textit{VLNVerse} protocol. In contrast, \ours is the only method equipped with low-level control and operates under realistic physical constraints. Nevertheless, as shown in~\Cref{tab:baseline}, \ours achieves the highest SR and the lowest reported NE on both \textit{VLNVerse-seen} and \textit{VLNVerse-unseen}, while achieving SPL and OSR comparable to the state-of-the-art RDP baseline. These results highlight the potential of whole-body navigation methods in cluttered indoor environment traversal. The relatively lower SPL suggests reduced navigation efficiency, likely due to the conservative behavior of the low-level tracker, which may favor safer but less direct trajectories around obstacles.

\subsection{Cluttered Environment Traversal Performance}

\begin{table}[t]
\centering
\captionsetup{skip=4pt,justification=raggedright}
\begin{minipage}[t]{0.53\textwidth}
\vspace{0pt}
\centering
{\small
\setlength{\tabcolsep}{2.4pt}
\renewcommand{\arraystretch}{0.95}

\begin{tabular}{lcccc}
\toprule
\multirow{2}{*}{Methods} & \multicolumn{4}{c}{Augmented Val Unseen} \\
\cmidrule(lr){2-5}
& NE$\downarrow$ & SR$\uparrow$ & SPL$\uparrow$ & CR$\downarrow$ \\
\midrule

InternVLA-N1 \textit{(zero-shot)}$^\dagger$ & 5.34 & 26.67 & 11.92 & 19.03 \\
InternVLA-N1 \textit{(zero-shot)}$^\ddagger$ & \textbf{3.19} & 35.29 & 24.35 & 16.11 \\

\midrule
InternVLA-N1 + Unitree WBC & 4.57 & 35.00 & 17.38 & 17.49 \\
InternVLA-N1 + HumanoidPF & 4.04 & 41.88 & 29.49 & 15.81 \\
\midrule

\rowcolor{tango_red!25}

Ours
& 4.01 & \textbf{43.75} & \textbf{31.83} & \textbf{9.90} \\

\bottomrule
\end{tabular}

\par\vspace{0.25em}
\makebox[\linewidth][l]{%
\begin{minipage}{\linewidth}
\raggedright
$^\dagger$ Equipped with Unitree low-level controller and MPC.\\
$^\ddagger$ Tracked by HumanoidPF generalist policy
\end{minipage}%
}
}

\caption{\textbf{Cluttered environment evaluation. }We report performance comparison of \ours on \textit{augmented VLNVerse-unseen} against strong modular baselines. }
\label{tab:zero_shot_main}

\end{minipage}\hfill
\begin{minipage}[t]{0.45\textwidth}
\vspace{0pt}
\centering
{\fontsize{8.5}{10}\selectfont
\setlength{\tabcolsep}{0.8pt}
\renewcommand{\arraystretch}{1.12}
\clipbox{0pt 0pt 0pt 0pt}{%
\begin{tabularx}{\linewidth}{@{}l*{6}{>{\centering\arraybackslash}X}@{}}
\toprule
\multirow{2}{*}{Method}
& \multicolumn{2}{c}{\makebox[0pt]{\shortstack{Short-horizon\\2D Env.}}}
& \multicolumn{2}{c}{\makebox[0pt]{\shortstack{Long-horizon\\2D Env.}}}
& \multicolumn{2}{c}{\makebox[0pt]{\shortstack{Cluttered\\3D Env.}}} \\
\cmidrule(lr){2-3}\cmidrule(lr){4-5}\cmidrule(l){6-7}
& SR$\uparrow$ & Coll.$\downarrow$ & SR$\uparrow$ & Coll.$\downarrow$ & SR$\uparrow$ & Coll.$\downarrow$ \\
\midrule
InternVLA-N1$^\dagger$
& 11/15 & 1.40 & 6/15 & 3.47 & 6/15 & 1.93 \\
\addlinespace[2pt]
\rowcolor{tango_red!25}[\tabcolsep][\dimexpr\tabcolsep+0.2pt\relax]
Ours & \textbf{12/15} & \textbf{0.40} & \textbf{8/15} & \textbf{1.07} & \textbf{10/15} & \textbf{0.73} \\
\bottomrule
\end{tabularx}%
}%
\par\vspace{0.25em}
\raggedright
$^\dagger$ Equipped with Unitree WBC.\par
}
\caption{\textbf{Quantitative real-world navigation results.} Three scenes with five trials each give 15 trials per method and setting. SR: successful trials out of 15; Coll.: mean collisions per trial.}
\label{tab:real_world_quant}

\end{minipage}
\end{table}

\looseness=-1
We evaluate our method on \textit{augmented VLNVerse-unseen} scenes (\Cref{sec:data_augmentation}) to verify its obstacle-avoidance and spatial understanding capability in 3D environments. We compare our method with InternVLA-N1 in both zero-shot and fine-tuned settings, with two types of low-level executors: Unitree official RL controller~\cite{unitree_rl_gym} with a model predictive control (MPC) module, which executes basic movement according to planar velocity command; and HumanoidPF generalist policy~\cite{xue2026cat} which performs obstacle-avoidance motions in complex 3D scenes. 
For the fine-tuned setting, we fine-tune InternVLA-N1 in \textit{augmented VLNVerse-train} scenes, with visual input collected from either Unitree-controller or HumanoidPF motions (denoted as ``+Unitree WBC'' and ``+HumanoidPF''). 
During inference, Unitree-controller takes in 2D velocity commands interpreted from predicted trajectory; while HumanoidPF tracks waypoints sampled 1.2m ahead on the trajectory, both at 50Hz. 
Note that while HumanoidPF leverages LiDAR as additional input for 3D scene information, our method uses pure RGB input. 

To better quantify navigation safety in cluttered environments, we introduce Collision Rate (CR)~\cite{lin2025vlnverse}, defined as the percentage of evaluation episodes with at least one collision (\Cref{app:metrics}). As shown in~\Cref{tab:zero_shot_main}, \ours achieves the highest SR and SPL while maintaining the lowest CR across all methods. In detail, \ours reduces CR from 15.81\% to 9.90\% compared to the strongest baseline, despite relying solely on RGB observations, whereas the HumanoidPF-tracked baseline additionally has access to LiDAR-based geometric perception. Meanwhile, \ours improves SR by 1.87 percentage points and SPL by 2.34 points against the fine-tuned InternVLA-N1 + HumanoidPF baseline. These results support the benefit of end-to-end whole-body action generation over the evaluated modular approaches.

\subsection{Real-World Experiment}

\looseness=-1
We conduct real-world experiments to evaluate whether \ours can transfer from simulation to a physical humanoid platform without any real-world training. In particular, we focus on scenarios that require simultaneous language-guided navigation, obstacle avoidance, and whole-body motion adaptation, which jointly test the key capabilities targeted by our approach.

\looseness=-1
\textbf{Qualitative Experiments. }We evaluate \ours in four representative real-world scenarios: long-horizon navigation, side-stepping through a narrow pathway, bending down to avoid overhead obstacles, and stepping over obstacles on the ground. As shown in~\Cref{fig:real_world}, \ours demonstrates robust scene understanding and spatial traversal capability in all cases. The robot executes continuous whole-body motions while following natural-language instructions, adapts its body configuration to negotiate obstacles, and maintains progress toward the navigation goal, demonstrating zero-shot sim-to-real transfer in cluttered physical environments.

\looseness=-1
\textbf{Quantitative Experiments.} We compare \ours with the fine-tuned InternVLA-N1 + Unitree WBC baseline in three settings: short-horizon 2D navigation (one turn, approximately 10\,m), long-horizon 2D navigation (two or three turns, approximately 30\,m), and cluttered 3D navigation with one challenging obstacle. Each setting contains three scenes with five trials per scene, giving 15 trials per method per setting. We report successful trials and mean collisions per trial in~\Cref{tab:real_world_quant}; the latter is a collision count, distinct from the episode-level CR used in simulation. The cluttered-scene instructions explicitly specify the required traversal behavior, such as stepping over, side-stepping, or bending down.
As shown in the table, \ours scores the highest SR \reviewtext{across the three tested settings}, \reviewtext{with fewer collisions than the baseline}. These results support improved task completion and safer traversal across the three tested settings.

\subsection{Ablation Studies}

\textbf{Whole-body action representation.} We study the impact of action representation and low-level execution on the original \textit{VLNVerse-unseen} benchmark to validate our 29-DoF whole-body action space. We compare against InternVLA-N1 in a zero-shot setting and a planar variant of our method, denoted as \textit{Ours-2D}. Since planar trajectory prediction is an auxiliary objective in our formulation (\Cref{sec:main_method}), \textit{Ours-2D} isolates the effect of whole-body action generation. For both baselines, we evaluate under two execution settings: teleportation and physical execution using the Unitree low-level controller with MPC.

\clearpage
\begingroup
\makeatletter
\setlength{\@fptop}{0pt plus 1fil}
\setlength{\@fpbot}{0pt plus 1fil}
\makeatother
\begin{figure}[p]
  \centering
  \includegraphics[width=\textwidth]{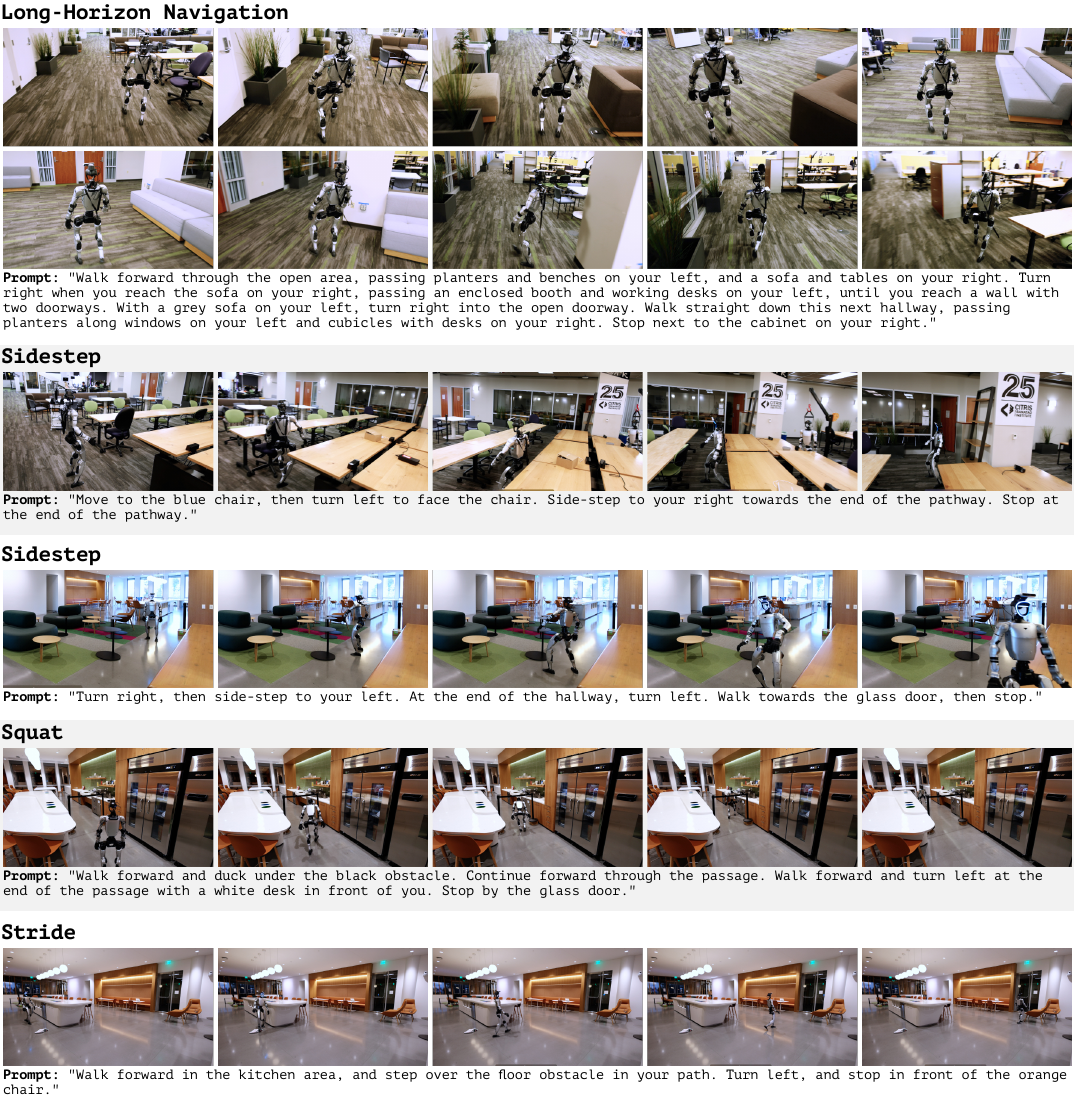}
  \caption{\textbf{Real-world deployment of \ours.}
  We show real-world qualitative results of \ours in long-horizon navigation, side-stepping through narrow pathways, bending down to avoid overhead obstacles, and stepping over obstacles on the ground, demonstrating zero-shot sim-to-real transfer without any real-world training.}
  \label{fig:real_world}
  \label{fig:trajectory_gallery}
\end{figure}
\clearpage
\endgroup

\newlength{\actionmethodwidth}
\newlength{\componentmethodwidth}
\newlength{\ablationcontrolwidth}
\newlength{\ablationmetricwidth}
\newlength{\actiontablewidth}
\newlength{\componenttablewidth}
\settowidth{\actionmethodwidth}{{\small InternVLA-N1$^\dagger$}}
\settowidth{\componentmethodwidth}{{\small SONIC $\rightarrow$ ScaleBFM}}
\settowidth{\ablationcontrolwidth}{{\small\shortstack{Low-level\\control}}}
\setlength{\ablationmetricwidth}{\dimexpr(\textwidth-\actionmethodwidth-\componentmethodwidth-\ablationcontrolwidth-36pt-14pt)/5\relax}
\setlength{\actiontablewidth}{\dimexpr\actionmethodwidth+\ablationcontrolwidth+2\ablationmetricwidth+18pt\relax}
\setlength{\componenttablewidth}{\dimexpr\componentmethodwidth+3\ablationmetricwidth+18pt\relax}
\begin{table}[t]
\centering
\captionsetup{skip=4pt,justification=raggedright}
\begin{minipage}[t]{\actiontablewidth}
\vspace{0pt}
\centering
{\small
\setlength{\tabcolsep}{3pt}
\renewcommand{\arraystretch}{1}
\clipbox{0pt 0pt 0pt 0pt}{%
\begin{tabular}{@{}l>{\centering\arraybackslash}p{\ablationcontrolwidth}*{2}{>{\centering\arraybackslash}p{\ablationmetricwidth}}@{}}
\toprule
\multirow{2}{*}{Method}
& \multirow{2}{*}{\shortstack{Low-level\\control}}
& \multicolumn{2}{c}{Val Unseen} \\
\cmidrule(l){3-4}
& & SR$\uparrow$ & SPL$\uparrow$ \\
\midrule
InternVLA-N1 & \xmark & 43.69 & 35.74 \\
Ours-2D & \xmark & 45.74 & 35.44 \\
\midrule
InternVLA-N1$^\dagger$ & \cmark & 42.37 & 22.50 \\
Ours-2D$^\dagger$ & \cmark & 26.67 & 8.34 \\
\rowcolor{tango_red!25}[\tabcolsep][\dimexpr\tabcolsep+0.2pt\relax]
Ours & \cmark & \textbf{52.89} & \textbf{40.18} \\
\bottomrule
\end{tabular}%
}%
}
\par\vspace{0.25em}
{\small\raggedright $^\dagger$ Equipped with Unitree low-level controller and MPC.\par}
\caption{\textbf{Ablation studies on action space.} We ablate the effect of action space and low-level control in \textit{VLNVerse-unseen}.}
\label{tab:ablation}

\end{minipage}\hfill
\begin{minipage}[t]{\componenttablewidth}
\vspace{0pt}
\centering
{\small
\setlength{\tabcolsep}{3pt}
\renewcommand{\arraystretch}{1}
\clipbox{0pt 0pt 0pt 0pt}{%
\begin{tabular}{@{}l*{3}{>{\centering\arraybackslash}p{\ablationmetricwidth}}@{}}
\toprule
\multirow{2}{*}{Method}
& \multicolumn{3}{c}{Augmented Val Unseen} \\
\cmidrule(l){2-4}
& SR$\uparrow$ & SPL$\uparrow$ & CR$\downarrow$ \\
\midrule
w/o RTC & 10.94 & 10.94 & 14.60 \\
w/o Motion Editing & 36.25 & \underline{30.36} & 20.60 \\
SONIC $\rightarrow$ ScaleBFM & \underline{40.94} & 29.65 & \textbf{9.10} \\
\midrule
\rowcolor{tango_red!25}[\tabcolsep][\dimexpr\tabcolsep+0.2pt\relax]
Ours & \textbf{43.75} & \textbf{31.83} & \underline{9.90} \\
\bottomrule
\end{tabular}%
}%
}
\caption{\textbf{Ablation studies on key components.} We evaluate RTC, motion editing, and the low-level WBC module on \textit{augmented VLNVerse-unseen}. The default configuration uses SONIC; the tracker variant uses ScaleBFM. SR and CR are percentages; SPL is scaled by 100.}
\label{tab:component_ablation}

\end{minipage}
\end{table}

As shown in~\Cref{tab:ablation}, both planar policies experience a substantial performance drop when moving from teleportation to physical execution, highlighting the difficulty of transferring conventional navigation policies to embodied settings. In contrast, \ours consistently outperforms all baselines and remains robust under low-level control constraints. The comparison between Ours-2D and \ours supports the benefit of whole-body action prediction. The zero-shot InternVLA-N1 rows provide additional reference points, but do not isolate action representation from training differences.

\textbf{Other Key Components.} We separately evaluate the design choice of \textit{motion editing, RTC, and low-level tracker} on \textit{augmented VLNVerse-unseen} (\Cref{tab:component_ablation}). Removing RTC reduces SR from 43.75\% to 10.94\%, a drop of 32.81 percentage points, and increases CR from 9.90\% to 14.60\%. This result supports the importance of maintaining motion continuity across action chunks during execution. Removing motion editing reduces SR to 36.25\% and increases CR to 20.60\%, indicating that obstacle-aware motion supervision contributes to collision avoidance. Replacing SONIC with ScaleBFM~\cite{zeng2026scaling} yields 40.94\% SR, 29.65 SPL, and 9.10\% CR, with all three metrics within three points of the default configuration. This comparison suggests compatibility with another general whole-body controller.

\setlength{\textfloatsep}{\experimentfloatsep}

\section{Conclusion}
This work introduces \oursb, to our best knowledge, the first whole-body vision-language navigation framework that directly predicts 29-DoF joint-space actions. We build up a diverse large-scale dataset covering not only indoor navigation patterns but also whole-body collision avoidance prior. Given language instruction, image observations and robot proprioception, we train a Qwen2.5VL-7B \cite{bai2025qwen25vl} based backbone with flow matching action expert. To match real-time execution on humanoid robot, training-time RTC \cite{black2025trainingtimertc} and high-frequency general tracker \cite{luo2025sonic} are leveraged. 

Our experiment results indicate both state-of-the-art vision-language navigation capability and decent traversal performance in cluttered environments. Our ablations support the benefits of whole-body action prediction, obstacle-aware motion editing, and RTC, and show compatibility with an alternative low-level tracker.

Overall, \oursb represents a meaningful step toward practical whole-body large planning models by demonstrating the feasibility of directly predicting 29-DoF actions. Building on \oursb as a foundation model, future work can further adapt and scale this framework toward more general loco-manipulation foundation models capable of tackling challenging tasks that require coordinated and active use of the robot’s entire body.

\textbf{Limitation. }Despite the promising results, the capability of low-level tracker shows up as a key constraint for further deployment in more complex environments, \textit{e.g.} walking up stairs. Another limitation lies in our vision input, which relies solely on RGB images. This may restrict the model’s ability to fully understand complex scenes, particularly in cluttered, visually ambiguous, or low-light environments, where depth camera and LiDAR are expected to help. We leave these for future work to explore.

\clearpage
\bibliographystyle{unsrt}
\bibliography{main}

\clearpage
\tableofcontents
\clearpage
\beginappendix{
\reviewchangesfalse
\crefalias{section}{appendix}
\crefalias{subsection}{appendix}
\crefalias{subsubsection}{appendix}

\section{Training Details}
For our training procedure, we emphasize rare whole-body behaviors through both data-level rebalancing and loss-level weighting. After converting rollouts into action chunks, we rebalance the training set by upsampling behaviorally important rows, including large turns, sideways motion, squat, stride, and mixed squat-stride segments; and especially, stride and sideways motions are boosted most strongly to compensate for their lower frequency and weaker action signal. We further apply motion-conditioned per-dimension loss weights on the action vector, assigning higher weights to the joint-position dimensions most associated with each behavior, such as hip, spine, and knee joints for squat, spine and shoulder joints for stride, their union for mixed motions, and shoulder plus body-frame motion dimensions for sideways motion. 
The model is trained with full-parameter tuning on 16 nodes of 8 $\times$ NVIDIA A100 GPUs for approximately 7 hours, totaling 896 A100 GPU hours.

\section{Deployment Details}
\label{app:impl_details}

The real-world system runs the VLA on an RTX PRO 6000 server and the SONIC tracker on the onboard Jetson Orin NX. Front and downward RGB observations from the RealSense D455 and D435i, together with proprioception, are transmitted to the server with approximately 20\,ms latency. The server performs VLA inference every 0.5\,s. Each execution segment contains 15 actions at 30\,Hz, which are resampled to 50\,Hz before tracking; the low-level control loop runs at approximately 200\,Hz. A web-based control panel provides navigation-instruction entry, observation and predicted-motion monitoring, and control signals to the VLA and WBC systems.

\section{Environment Augmentation and Motion Generation Details}
\subsection{Environment Augmentation Details}
\label{app:env_aug_details}

\begin{figure*}[!t]
  \centering
  \setlength{\tabcolsep}{2pt}
  \renewcommand{\arraystretch}{1.02}
  \resizebox{0.95\textwidth}{!}{
  \begin{tabular}{@{}c@{\hspace{4pt}}ccccc@{}}
    \rotatebox{90}{Stride} &
    \includegraphics[width=0.18\linewidth]{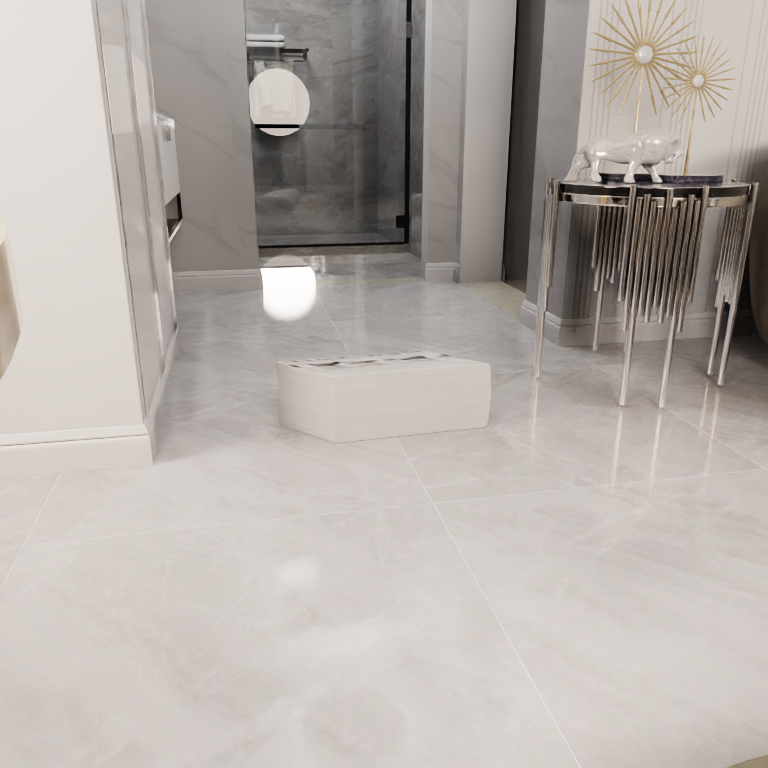} &
    \includegraphics[width=0.18\linewidth]{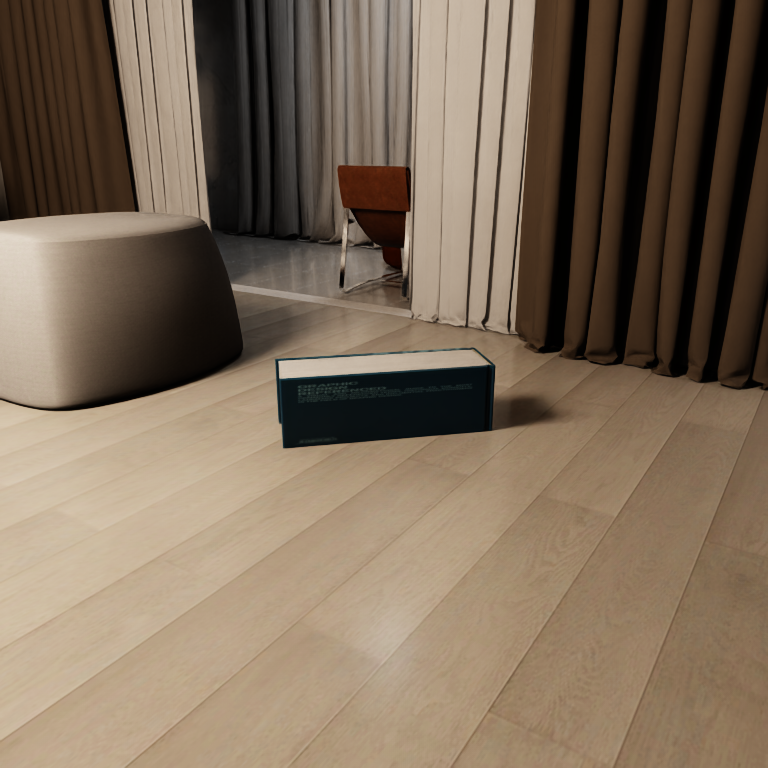} &
    \includegraphics[width=0.18\linewidth]{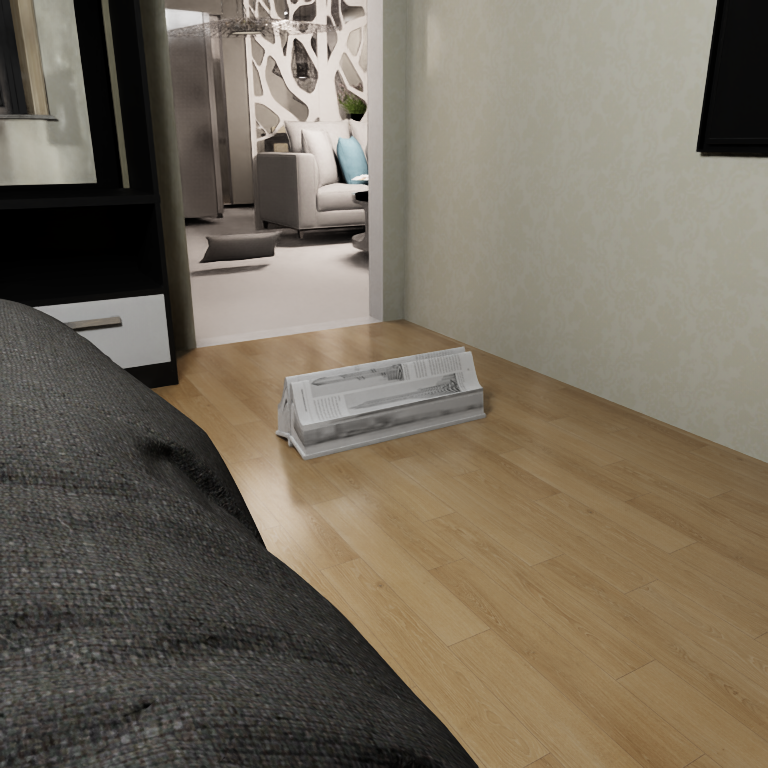} &
    \includegraphics[width=0.18\linewidth]{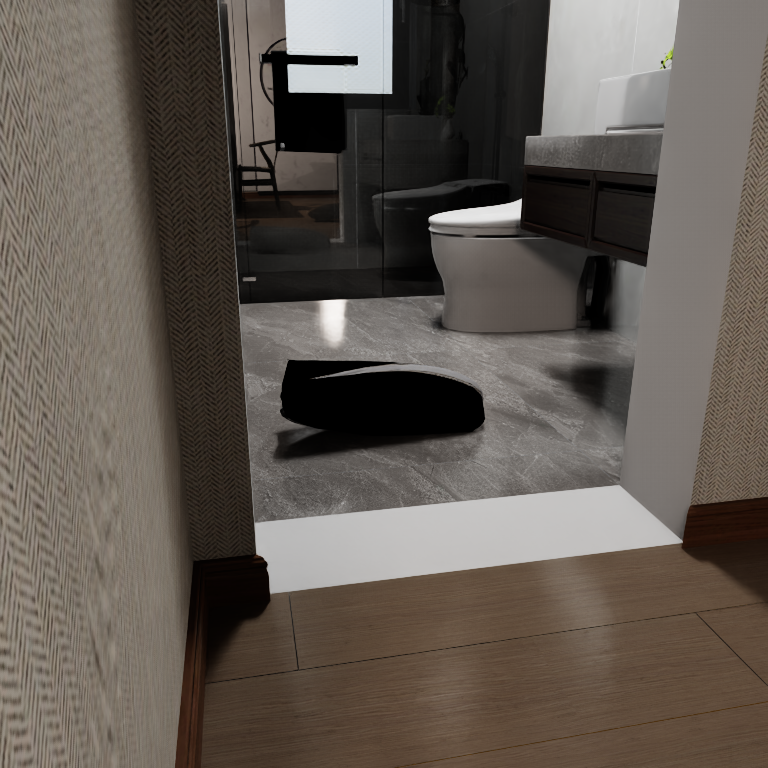} &
    \includegraphics[width=0.18\linewidth]{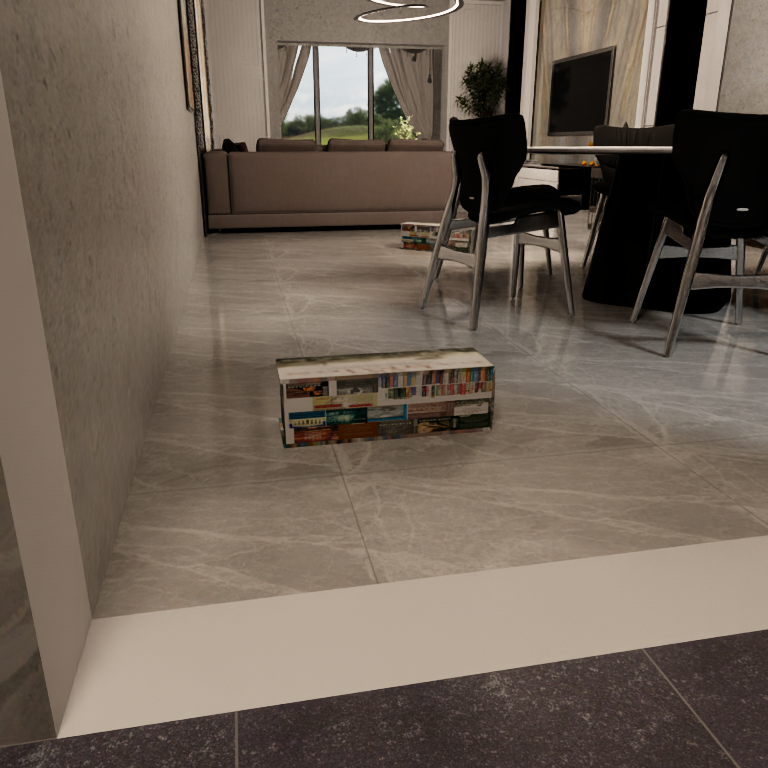} \\[3pt]

    \rotatebox{90}{Sidle} &
    \includegraphics[width=0.18\linewidth]{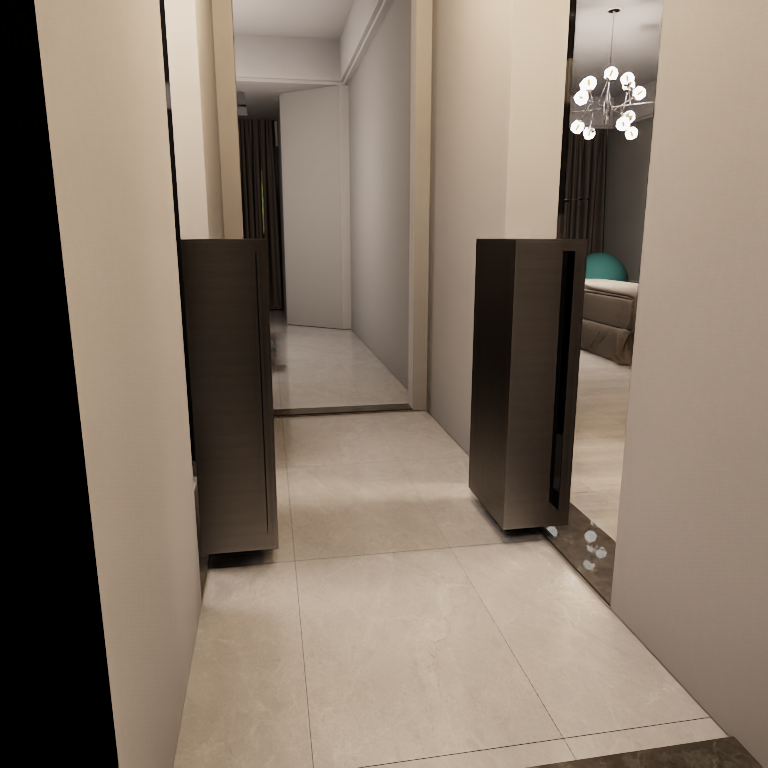} &
    \includegraphics[width=0.18\linewidth]{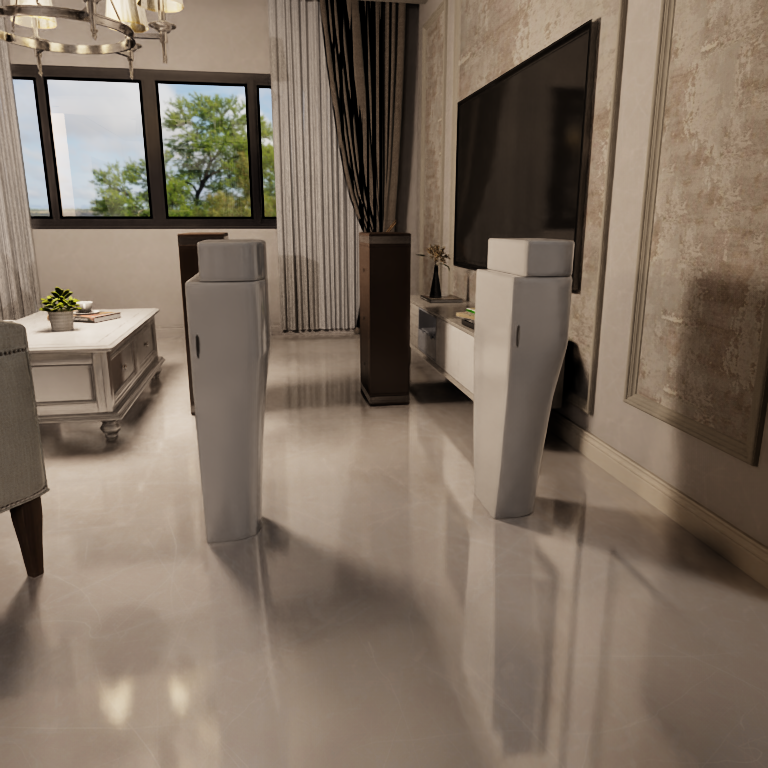} &
    \includegraphics[width=0.18\linewidth]{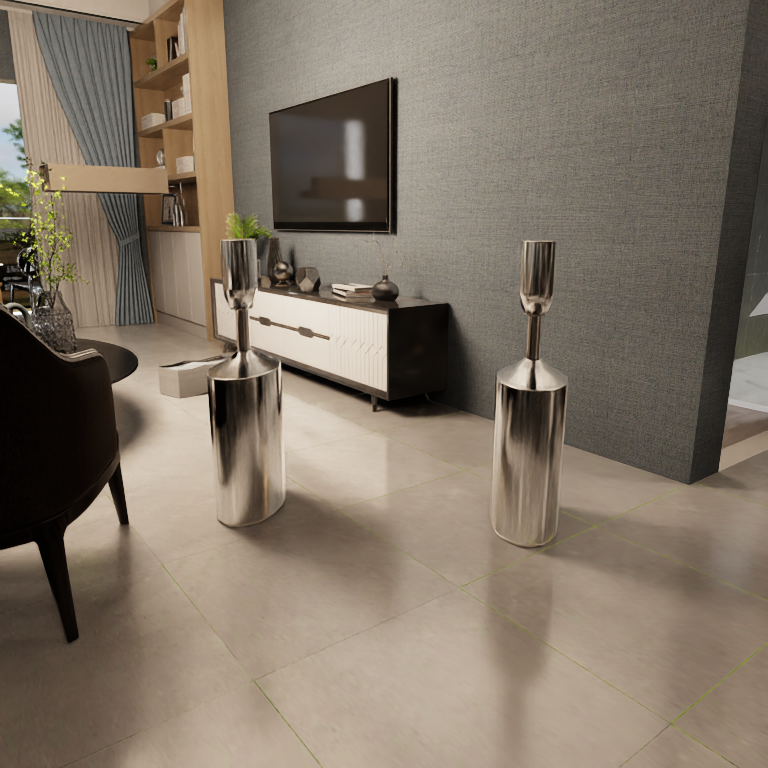} &
    \includegraphics[width=0.18\linewidth]{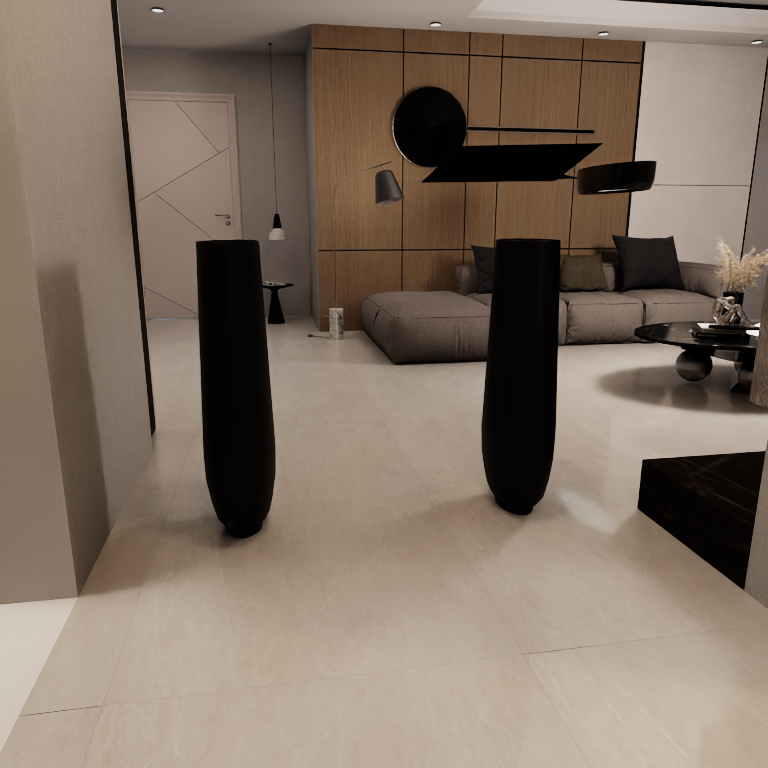} &
    \includegraphics[width=0.18\linewidth]{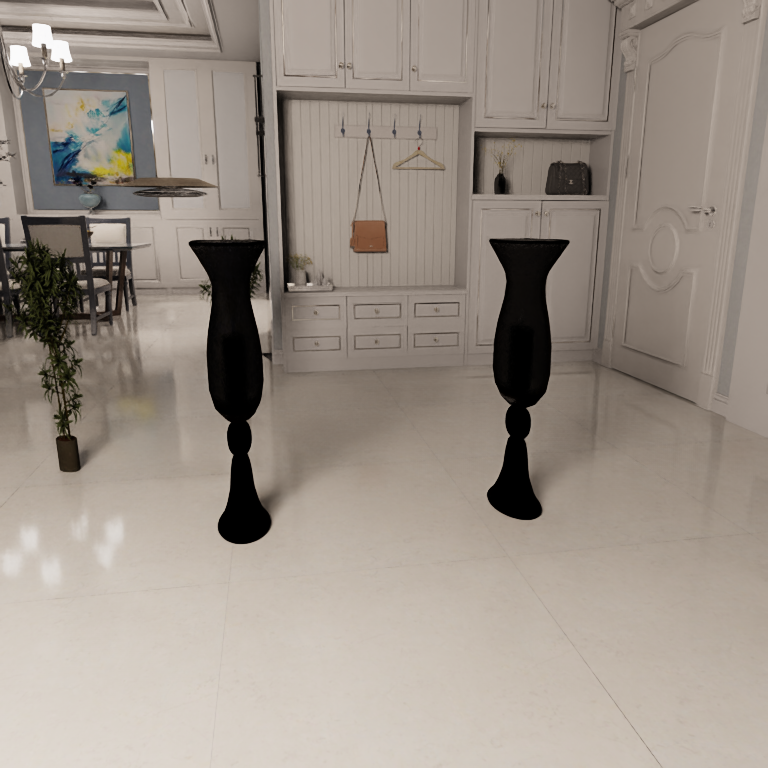} \\[3pt]

    \rotatebox{90}{Squat} &
    \includegraphics[width=0.18\linewidth]{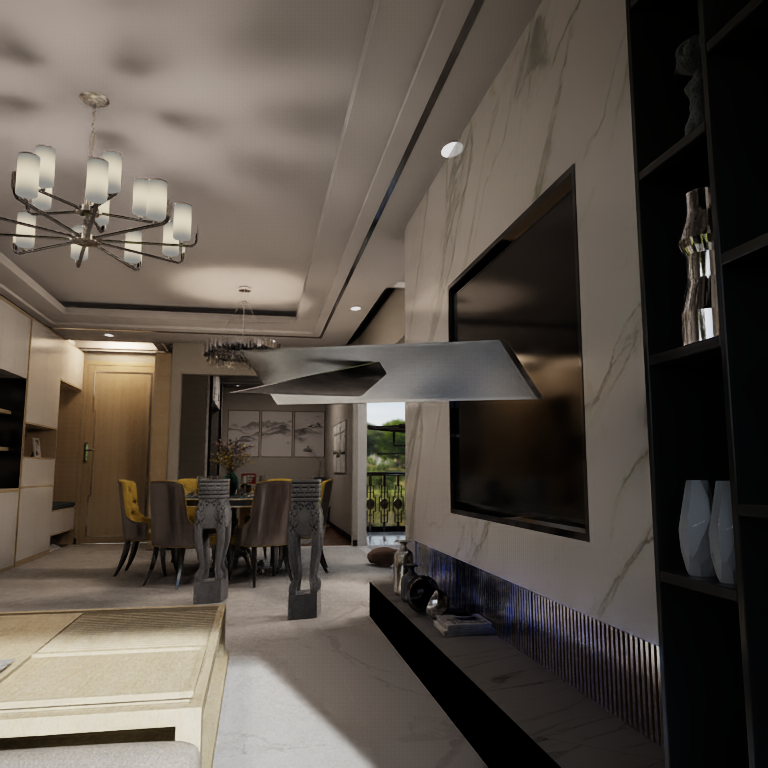} &
    \includegraphics[width=0.18\linewidth]{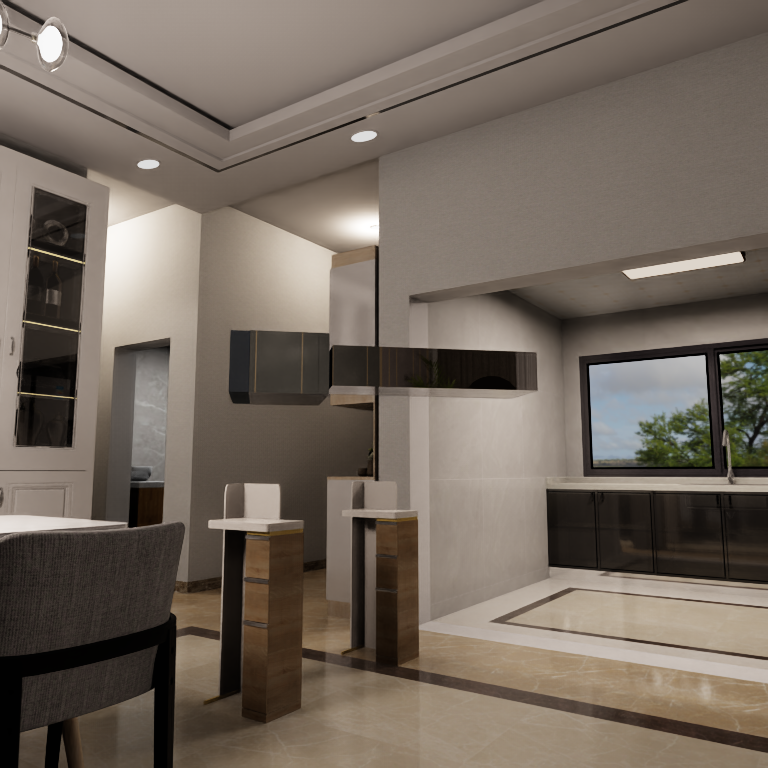} &
    \includegraphics[width=0.18\linewidth]{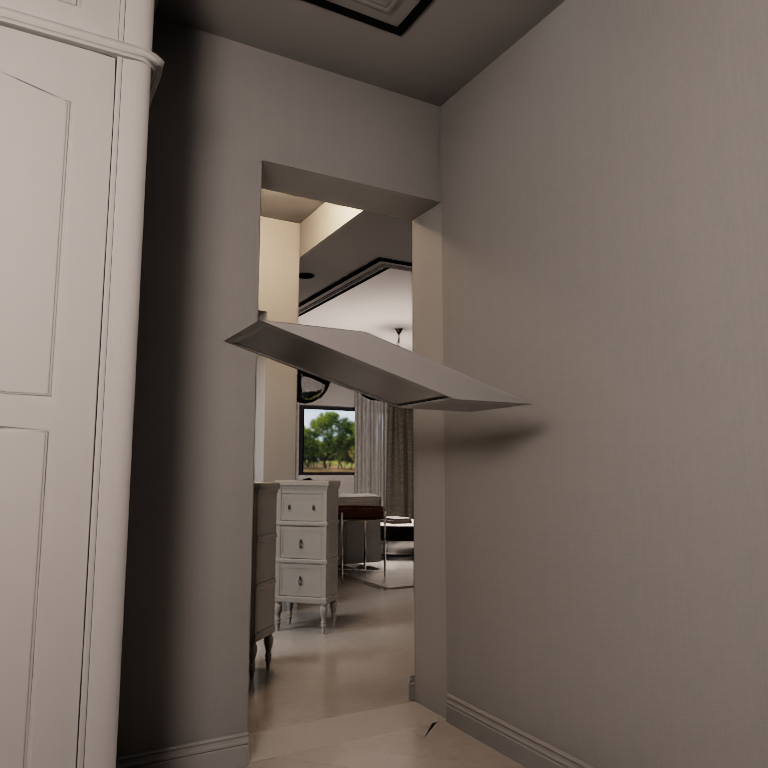} &
    \includegraphics[width=0.18\linewidth]{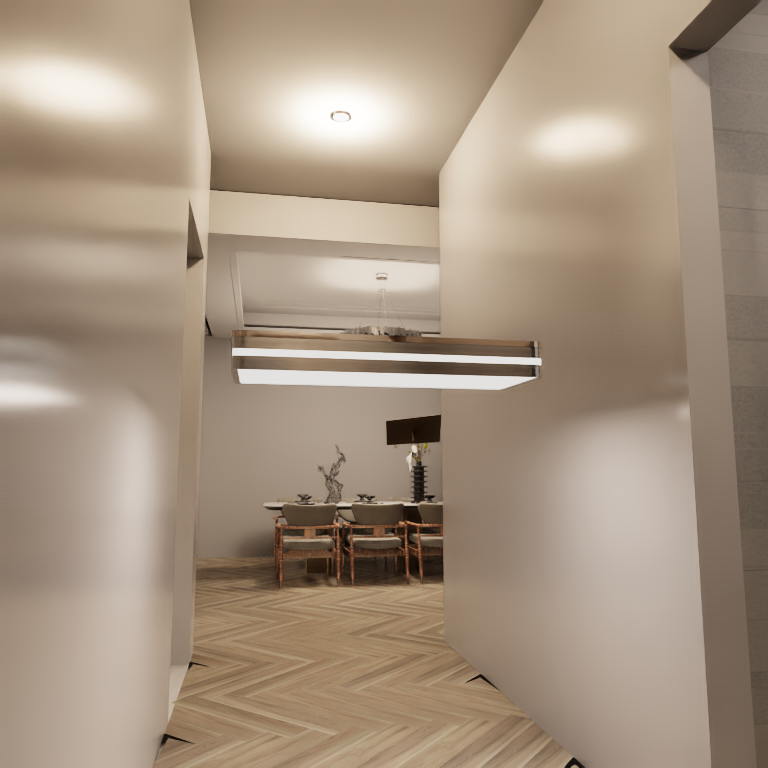} &
    \includegraphics[width=0.18\linewidth]{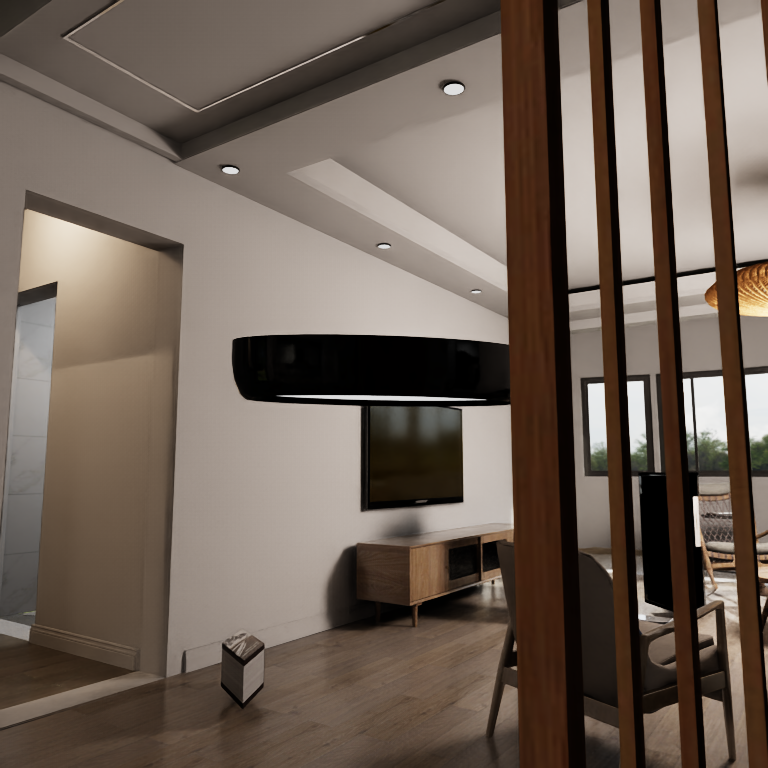}
  \end{tabular}
  }
  \caption{
  \textbf{Examples of targeted environment augmentation.}
  Examples of \textit{augmented VLNVerse} scenes with three obstacle types: low obstacles for stepping over, side obstacles for sidling through narrow passages, and overhead obstacles for squatting.
  }
  \label{fig:targeted-aug-gallery}
\end{figure*}

To better evaluate whole-body planning, \reviewtext{we augment \textit{VLNVerse} and \textit{SAGE-3D} scenes} with trajectory-conditioned obstacles rather than randomly sampled clutter. Given an original scene, its A* navigation trajectories, and the corresponding occupancy map extracted from the scene, we insert obstacles along the paths at the locations where whole-body behaviors are likely required. We consider three targeted interaction types: \textit{stride}, where a low obstacle is placed on the path to encourage stepping over; \textit{sidle}, where a pair of side obstacles forms a narrow passage; and \textit{squat}, where an overhead or torso-height obstacle encourages ducking or lowering the body.

For each scene, candidate placements are sampled along valid A* trajectories while avoiding the start and goal regions. Each candidate is aligned with the local path direction so that the inserted obstacle naturally interacts with the robot's intended route. The obstacle assets are selected according to the target behavior: low objects such as rugs, cushions, or stools for \textit{stride}; chairs, plants, shelves, or similar objects for \textit{sidle}; and ceiling lights, lamps, curtains, or wall-mounted objects for \textit{squat}. This design makes the augmented scenes semantically plausible while explicitly inducing whole-body navigation challenges.

To preserve scene validity, each placement is checked against the occupancy map and all available trajectories in the same scene. In particular, obstacles are prevented from blocking unrelated paths or appearing too close to trajectory endpoints, while the source trajectory is allowed to be affected by the inserted obstacle. After placement, the augmented obstacles are also written into the occupancy representation so that downstream planners observe the same geometry as the policy. \reviewtext{From the original 263 \textit{VLNVerse} and 1,000 \textit{SAGE-3D} scenes, Gemini 2.5 Flash filtering removes low-quality scenes, such as those with uneven ground or incomplete geometry. We retain 205 \textit{VLNVerse} and 373 \textit{SAGE-3D} scenes. Across the augmented scenes, the proportions of stride, sidle, and squat obstacles are 44\%, 15\%, and 41\%, respectively. Obstacle poses are randomly sampled within traversable ranges rather than fixed for each category.} Please refer to \Cref{fig:targeted-aug-gallery} for a visualization of the augmented scenes.

To better align instructions and action labels to facilitate training, we further augment the language prompts from \textit{VLNVerse} with the knowledge of added obstacles and their corresponding target behavior types via Gemini 2.5 Flash~\cite{comanici2025gemini25}. In practice, including collision-avoidance prompts helps convergence and the learning of obstacle-avoidance behaviors. In the RGB-only real-world setting, these prompts provide complementary cues to visible obstacles. This is an empirical observation, rather than a controlled language ablation, and the reported cluttered real-world trials use behavior-explicit instructions.

\clearpage
\begin{table}[t]
\centering
\begin{reviewblock}
\small
\setlength{\tabcolsep}{12pt}
\begin{tabular}{lrrrr}
\toprule
\multirow{2}{*}{Dataset} & \multirow{2}{*}{Trajectories}
& \multicolumn{3}{c}{GPU-hours} \\
\cmidrule(lr){3-5}
& & PET & Rendering & Total \\
\midrule
VLNVerse (original) & 3,478 & 3 & 8 & 11 \\
VLNVerse (augmented) & 3,784 & 6 & 17 & 23 \\
SAGE-3D (original) & 28,678 & 20 & 45 & 65 \\
SAGE-3D (augmented) & 28,693 & 57 & 55 & 112 \\
\midrule
Total & \textbf{64,633} & \textbf{86} & \textbf{125} & \textbf{211} \\
\bottomrule
\end{tabular}
\end{reviewblock}
\caption{\reviewtext{\textbf{Dataset scale and generation cost.} Costs are measured in RTX PRO 6000 GPU-hours and cover PET motion generation and rendering for the original and augmented scenes.}}
\label{tab:data_generation_cost}
\end{table}

\reviewtext{\textbf{Dataset scale and generation cost.} \Cref{tab:data_generation_cost} reports the trajectory counts and GPU-hours for the original and augmented versions of both datasets. Generating 64,633 trajectories requires 86 GPU-hours for PET and 125 GPU-hours for rendering, totaling 211 RTX PRO 6000 GPU-hours. These costs cover trajectory generation and rendering, not VLA training.}

\subsection{Motion Generation Details}
\label{app:motion_generation}

This section describes the Plan, Edit, Track (PET) pipeline used to generate the trajectories that supervise \textsc{TANGO}. Given an augmented indoor scene and a start--goal pair, PET produces a 29-DoF whole-body reference motion, then re-renders egocentric observations and regenerates language instructions. The pipeline consists of three stages: Plan, Edit (which contains the motion editing deferred from the main text), and Track. 

\textbf{\textit{Plan}: Path Planning and Reference-Gait Synthesis}
\label{app:plan}
Given the start $s$, the goal $g$, and the scene occupancy map, which already includes the augmented obstacles , we plan a planar path with A$^\star$ on the 2D floor grid. To favor safer routes with whole-body clearance, we add an obstacle-aware term to the step cost. This term is computed from the 2D signed distance to occupied cells, $\Phi_{2\mathrm{D}}(\mathbf{x})$:
\begin{equation}
  c(\mathbf{x}) \;=\; c_{\text{step}}
  \;+\; \lambda\,\exp\!\big(-\Phi_{2\mathrm{D}}(\mathbf{x}) / d_0\big),
\end{equation}
so cells near obstacles receive a soft penalty that decays with clearance $d_0$, while the admissible Euclidean-to-goal heuristic preserves optimality under this cost. The output is a planar polyline $\mathcal{P}=\{(x_k,y_k)\}_{k=1}^{K}$ with a tangent heading $\psi_k=\operatorname{atan2}(y_{k+1}-y_k,\,x_{k+1}-x_k)$.

To enable side walking through narrow passages, we used a rule-based yaw rewriter. Along $\mathcal{P}$, we detect narrow passages by probing the lateral free width $w_\perp(s)$ on both sides of the path at arc length $s$. When $w_\perp(s)$ falls below a threshold $w_{\min}$, the body heading is rewritten to be $90^\circ$ offset from the travel tangent, $\psi^{\text{body}}_k = \psi_k \pm \tfrac{\pi}{2}$, with the sign chosen so that the leading shoulder remains on the wider side. Smooth $\pm 90^\circ$ transitions into and out of the passage produce a crab-walking, or side-stepping, gait in which the direction of translation and the facing direction are decoupled.

The rewritten planar trajectory $\{(x_k,y_k,\psi^{\text{body}}_k)\}$ is converted into a stream of base velocity commands $\mathbf{u}_k=(v_x,v_y,\omega)_k$, where the forward and lateral linear velocities and yaw rates are obtained by finite-differencing the path under the rewritten heading. These commands are fed to \textsc{SONIC}~\cite{luo2025sonic}, a motion-tracking model trained on $\sim$800 hours of high-quality motion-capture data, which produces a natural whole-body gait. A PD controller corrects drift between the realized base pose and $\mathcal{P}$, yielding a dynamically feasible reference walk $\mathcal{M}^{\text{ref}}=\{\mathbf{q}^{\text{ref}}_t\}_{t=1}^{T}$ with $\mathbf{q}^{\text{ref}}_t\in\mathbb{R}^{29}$ joint angles plus the floating-base pose. This gait is smooth, but above the floor it remains obstacle-agnostic: it follows the planar route and side-walks through narrow passages, but it does not yet step over, duck under, or otherwise adapt the body to 3D obstacle geometry. These adaptations are introduced in the Edit stage.

\textbf{\textit{Edit}: Whole-Body Obstacle-Avoidance Editing}
\label{app:edit}
The Edit stage transforms the reference motion $\mathcal{M}^{\text{ref}}$ into a whole-body obstacle-avoidance motion. The reference gait is kept as a locomotion prior: its timing, gait phase, and motion style are preserved, and edits are applied locally only where the path corridor contains obstacles. At each frame, we solve a whole-body inverse-kinematics (IK) problem with four objectives: reference posture tracking, foot contact and landing targets, a center-of-mass (CoM) target, and potential-field (PF) link forces. These objectives cover the three behavior families introduced above. A potential-field crouch controller handles squat obstacles, a gait-adaptation module handles stride obstacles, and crab-walking from the Plan stage, together with lateral PF guidance, handles sidle passages.

\label{app:edit_corridor}
The scene voxel occupancy is converted into a 3D signed distance field $\Phi(\mathbf{p})$ using the fast marching method, which provides distance values and gradients $\nabla\Phi$ throughout the workspace. Thin structural obstacles, such as bars, are also represented as exact oriented boxes with an analytic SDF, so that clearance queries remain continuous. Editing is restricted to a spatial corridor, defined as a tube of half-width $w_{\text{corr}}$ ($\approx0.5$\,m) around the planned path. Every field sample whose horizontal position lies outside the tube is treated as free space,
\begin{equation}
  \tilde\Phi(\mathbf{p}) =
  \begin{cases}
    \Phi(\mathbf{p}), & \operatorname{dist}_{xy}(\mathbf{p},\mathcal{P}) \le w_{\text{corr}},\\[2pt]
    +\infty, & \text{otherwise,}
  \end{cases}
\end{equation}
and the guidance vector outside the tube is zeroed. As a result, the editor ignores clutter that A$^\star$ has already routed around, and only on-path obstacles that the robot must negotiate affect the edit.

\label{app:edit_pf}
Following HumanoidPF~\cite{xue2026cat}, we use a sampled repulsive field for avoidance and pass this guidance to the IK through SoftMimic-style pseudo-forces~\cite{margolis2025softmimic}. Instead of imposing hard collision constraints, each force is converted into a bounded soft displacement target that the whole-body IK balances against posture and stability terms. We compute a guidance field $\mathbf{g}(\mathbf{p})$ by combining the path tangent $\hat{\boldsymbol\tau}$, which encourages forward progress, with the SDF repulsion $\nabla\tilde\Phi$, which encourages obstacle avoidance. The weights are chosen so that repulsion dominates near obstacles:
\begin{equation}
  \mathbf{g}(\mathbf{p}) \;=\;
  \hat{\boldsymbol\tau}(\mathbf{p})
  \;+\; \beta\,e^{-\tilde\Phi(\mathbf{p})/\sigma}\,
  \frac{\nabla\tilde\Phi(\mathbf{p})}{\lVert\nabla\tilde\Phi(\mathbf{p})\rVert}.
\end{equation}
We apply forces to a set of key body links $\mathcal{L}$, including the shoulders, elbows, wrists, torso, and optionally the knees and pelvis. Feet and ankles are excluded so that the contact pattern is still determined by the reference gait and the gait-adaptation module. For each link $\ell\in\mathcal{L}$ at world position $\mathbf{p}_\ell$, we sample a repulsive force $\mathbf{f}_\ell = F_{\max}\,\mathbf{g}(\mathbf{p}_\ell)$ and convert it into a displacement under a link stiffness $\kappa$, clipped to a per-link cap $\delta_{\max}$:
\begin{equation}
  \Delta\mathbf{p}_\ell
  = \operatorname{clip}\!\Big(\tfrac{1}{\kappa}\,\mathbf{f}_\ell,\;\delta_{\max}\Big),
  \qquad
  \tilde{\mathbf{f}}_\ell = \kappa\,\Delta\mathbf{p}_\ell .
\end{equation}
The bounded pseudo-forces $\{\tilde{\mathbf{f}}_\ell\}$ are added as soft tasks to the whole-body IK and are temporally low-pass filtered to remove discontinuities caused by field changes at gait transitions. For squat obstacles, we keep only the vertical component of each force, so the field provides a body-lowering cue for ducking rather than a lateral push. For sidle passages, the horizontal components push the trunk and arms away from the side walls in coordination with the crab-walking heading. The per-link forces can also be aggregated into a common CoM offset, with only the residual deviation applied to each link, so that the body moves away from obstacles as a coherent whole rather than having each limb react independently.

\label{app:edit_crouch}
Two additional terms make ducking under squat obstacles start early and remain stable. The look-ahead term addresses the fact that the raw field can stay near zero until the body is already under a ceiling. We probe $\tilde\Phi$ ahead of the upper body at several head and shoulder heights and at several forward distances. If the minimum probe lies within a margin $\eta$ of the geometry, a downward force proportional to the violation is added, so the body begins to lower before reaching the obstacle. The virtual head barrier protects a point $\mathbf{p}_{\text{head}}=\mathbf{p}_{\text{torso}}+h\hat z$ with clearance $b=\tilde\Phi(\mathbf{p}_{\text{head}})-r$. When $b<b_m$, it adds a strictly downward force $k_s(b_m-b) + k_h\max(0,-b)$, corresponding to soft and hard barrier components, clipped to a maximum trunk displacement. The resulting downward displacement of the upper body is converted into a coordinated squat by coupling the crouch activation $\alpha=\operatorname{clip}(\Delta z/\delta_{\max},0,1)$ to posture: a forward waist-pitch target $\theta^{\text{waist}}\!\leftarrow\!\theta^{\text{waist}}+\alpha\,\theta_0$, a proportional hip-pitch bias, and a downward shift of the CoM target $\Delta z_{\text{CoM}}=\alpha\,c_z$. This CoM shift lets the legs lower the body rather than letting the CoM task pull the body back upright. The implementation supports three coordinated presets: waist-only, which bends at the waist while the tracker controls the legs; com-drop, which lowers the body vertically; and full-squat, which combines waist motion, pelvis tilt, hip flexion, knee bending, ankle dorsiflexion, and CoM lowering. During an active crouch, the upper-body reference-tracking terms are relaxed so that the field can reshape the trunk instead of competing with the upright reference.

\label{app:edit_gait}
For stride obstacles, upper-body guidance is not enough because the swing feet must step over the object. The gait-adaptation module retargets footsteps online while preserving the phase and timing of the reference gait. At each foot lift-off, the base is projected onto the planned path, a short arc-length look-ahead is queried, and the landing target is specified in the path-yaw frame. This target is clamped to a maximum displacement from the reference touchdown, which prevents the IK solver from pursuing unreachable foot placements. A footprint-aware SDF scan detects a low obstacle ahead of the swing foot and returns the interval $[d_{\text{near}}, d_{\text{far}}]$ spanned by the obstacle, together with its top height $z_{\text{top}}$. We classify obstacles by height: only $z_{\text{top}}\le z_{\text{bar}}$, corresponding to a steppable bar, triggers a step-over motion. Taller geometry is treated as a wall or ceiling and is handled by the lateral and upper-body layers, so the robot does not attempt to step over a wall. For a steppable bar, the landing target is pushed past the far edge by a margin, $d^{\text{land}} = d_{\text{far}} + m$. The landing objective is evaluated with respect to this target rather than the reference touchdown on the bar. This both clears the obstacle and compensates for the forward-reach undershoot of the downstream tracker, keeping the realized landing past the bar.

The swing trajectory is formed by warping the reference foot trajectory from motion capture or RL data onto the retargeted endpoints and adding a vertical clearance arc. This keeps the natural swing shape and leaves ordinary steps unchanged. With swing phase $u\in[0,1]$, after smoothstep interpolation to $\bar u$, and reference foot path $\mathbf{p}^{\text{ref}}(u)$,
\begin{equation}
  \mathbf{p}^{\text{swing}}(u) =
  \mathbf{p}^{\text{ref}}(u)
  + (1-\bar u)\,\Delta\mathbf{p}_{\text{lift}}
  + \bar u\,\Delta\mathbf{p}_{\text{land}}
  + h_{\text{arc}}\,\rho(u)\,\hat z,
\end{equation}
where $\Delta\mathbf{p}_{\text{lift}},\Delta\mathbf{p}_{\text{land}}$ are the offsets between the retargeted and reference endpoints, and $\rho(u)$ is an asymmetric clearance profile with a fast quarter-sine rise to an early peak followed by a quarter-cosine fall. The asymmetric profile keeps the foot high early in the swing, which is needed for the trailing leg that crosses the bar shortly after lift-off. A symmetric $\sin(\pi u)$ arc would be too low at this stage. The peak height $h_{\text{arc}}$ is solved for each step so that the full foot footprint clears the obstacle box at every sampled phase,
\begin{equation}
  \begin{aligned}
    z_{\text{ground}} + h_{\text{arc}}\,\rho(u)
    &\ge z_{\text{top}}(u) + \epsilon, \\
    &\forall\,u : \text{footprint}(u)\cap\text{obstacle}\neq\varnothing,
  \end{aligned}
\end{equation}
using the same profile $\rho$ in both the solver and the executed arc, so that the planned and executed motions remain consistent. A swing-knee bend bias encourages knee flexion rather than straight-leg extension for obstacle clearance. When several low obstacles appear in sequence, a short foothold sequence places natural-length steps through the gaps instead of requiring a single long step.

Because the foot retargeter moves the support polygon while the reference pelvis still follows the original gait, the body can lag behind the planned support. During a forward step-over, this mismatch can make the body lean in place rather than translate forward. Two small offsets, smoothed with EMA, correct this effect: root redirection shifts the pelvis reference toward the planned support, and CoM redirection shifts the IK CoM target accordingly. During a bar crossing, these offsets switch to a low-latency update rate so that the base translates with the crossing leg. This produces a forward step-over rather than an in-place lunge, with a ramped transition to avoid abrupt motion.

Each frame is solved in two passes. The first pass performs a lower-body projection that edits the reference to match the foot, support, and CoM targets. The second pass performs full compliant whole-body IK, which tracks the edited reference while applying the PF link forces. Strong foot-anchor costs are used so that the feet behave as contacts and posture edits do not induce foot sliding. The output is the edited 29-DoF whole-body trajectory $\mathcal{M}^{\text{edit}}$.

\textbf{\textit{Track}: Feasibility Restoration and Collision Filtering}
\label{app:track}
The edited trajectory $\mathcal{M}^{\text{edit}}$ is kinematic and may violate dynamic feasibility, for example through overly fast swings or aggressive crouches. We restore feasibility by passing it through the \textsc{SONIC} tracker~\cite{luo2025sonic} in closed loop. The edited whole-body reference, including joint positions, joint velocities, and base orientation, is streamed frame by frame to the RL tracking policy, which tracks it under physics in MuJoCo and outputs a dynamically feasible motion $\mathcal{M}^{\text{track}}$. The same tracker is used at deployment (system-0), making the synthesized supervision consistent with execution. The known forward undershoot of the tracker is also why the Edit stage uses generous step-over landing margins.

\section{Experiment Metrics}
\label{app:metrics}

We evaluate navigation performance with standard vision-language navigation
(VLN) metrics, including Success Rate (SR), Success weighted by Path Length
(SPL), Navigation Error (NE), and Oracle Success Rate (OSR). For cluttered-scene
traversal, we additionally report Collision Rate (CR) to quantify physical
safety.

Let $\mathcal{E}$ denote the evaluation set and $|\mathcal{E}|$ its number of
episodes. For each episode $e \in \mathcal{E}$, the policy receives a language
instruction $\ell_e$ and sequential observations as defined in
\Cref{sec:WBVLN}, and executes whole-body action chunks through the low-level
tracker. The resulting robot-base trajectory is denoted as
\begin{equation}
    \Gamma_e = (\mathbf{x}_{e,0}, \mathbf{x}_{e,1}, \ldots,
    \mathbf{x}_{e,K_e}),
\end{equation}
where $\mathbf{x}_{e,0}$ is the start position, $\mathbf{x}_{e,K_e}$ is the
final stopping position, and $K_e$ is the number of sampled trajectory points in
episode $e$. Let $\mathbf{g}_e$ denote the target goal position, and let
$d_{\mathcal{G}}(\cdot,\cdot)$ denote geodesic distance in the navigation
environment. We use $\delta$ as the success threshold, set to $3$ meters
following standard VLN evaluation. We further denote the executed path length as
\begin{equation}
    P_e = \sum_{k=1}^{K_e}
    \left\| \mathbf{x}_{e,k} - \mathbf{x}_{e,k-1} \right\|_2,
\end{equation}
and the shortest-path distance from the start position to the goal as
\begin{equation}
    L_e = d_{\mathcal{G}}(\mathbf{x}_{e,0}, \mathbf{g}_e).
\end{equation}

\paragraph{Navigation Error (NE).}
Navigation Error measures the average geodesic distance between the final
stopping position and the target goal:
\begin{equation}
    \mathrm{NE}
    =
    \frac{1}{|\mathcal{E}|}
    \sum_{e \in \mathcal{E}}
    d_{\mathcal{G}}(\mathbf{x}_{e,K_e}, \mathbf{g}_e).
\end{equation}
Lower NE indicates that the agent stops closer to the target goal.

\paragraph{Success Rate (SR).}
An episode is considered successful if the final stopping position is within the
success threshold $\delta$ of the target goal. The success indicator is
\begin{equation}
    S_e =
    \mathbf{1}
    \left[
    d_{\mathcal{G}}(\mathbf{x}_{e,K_e}, \mathbf{g}_e) \leq \delta
    \right],
\end{equation}
and SR is computed as
\begin{equation}
    \mathrm{SR}
    =
    \frac{1}{|\mathcal{E}|}
    \sum_{e \in \mathcal{E}}
    S_e.
\end{equation}
\reviewtext{In simulation tables, SR is reported as a percentage. In~\Cref{tab:real_world_quant}, SR is shown as successful trials out of 15.}

\paragraph{Oracle Success Rate (OSR).}
Following common VLN evaluation, Oracle Success Rate measures whether the
executed trajectory ever enters the goal region, regardless of the final
stopping position. The oracle success indicator is
\begin{equation}
    O_e =
    \mathbf{1}
    \left[
    \min_{0 \leq k \leq K_e}
    d_{\mathcal{G}}(\mathbf{x}_{e,k}, \mathbf{g}_e)
    \leq \delta
    \right],
\end{equation}
and OSR is computed as
\begin{equation}
    \mathrm{OSR}
    =
    \frac{1}{|\mathcal{E}|}
    \sum_{e \in \mathcal{E}}
    O_e.
\end{equation}
OSR therefore evaluates whether the trajectory reaches the goal neighborhood at
least once, rather than whether the agent stops there.

\paragraph{Success weighted by Path Length (SPL).}
SPL jointly measures task completion and path efficiency:
\begin{equation}
    \mathrm{SPL}
    =
    \frac{1}{|\mathcal{E}|}
    \sum_{e \in \mathcal{E}}
    S_e
    \frac{L_e}{\max(P_e, L_e)}.
\end{equation}
Failed episodes contribute zero to SPL, while successful but unnecessarily long
trajectories are penalized by the path-length ratio.

\paragraph{Collision Rate (CR).}
For cluttered-scene traversal, we report Collision Rate as an episode-level
safety metric. Let
\begin{equation}
    C_e =
    \mathbf{1}
    \left[
    \Gamma_e \text{ results in at least one collision}
    \right].
\end{equation}
CR is then defined as
\begin{equation}
    \mathrm{CR}
    =
    \frac{1}{|\mathcal{E}|}
    \sum_{e \in \mathcal{E}}
    C_e.
\end{equation}
Lower CR indicates safer whole-body traversal in cluttered environments. In our
tables, CR is reported as a percentage.

\reviewtext{\paragraph{Mean collisions in real-world trials.} For each real-world setting, let $n_e$ be the number of collisions recorded in trial $e$. We report $\mathrm{Coll.}=\frac{1}{15}\sum_{e=1}^{15}n_e$. Unlike CR, this metric counts multiple collisions within a trial and is not a percentage.}

}

\end{document}